\documentclass[journal]{IEEEtran}
\usepackage[nocompress]{cite}
\usepackage{times}
\usepackage{epsfig}
\usepackage{graphicx}
\usepackage{subfigure}
\usepackage{amsmath}
\usepackage{lipsum}
\usepackage{tabularx}
\usepackage{threeparttable}
\usepackage{multirow}
\usepackage{amssymb}
\usepackage{xcolor}
\usepackage{colortbl,booktabs}
\usepackage{ragged2e}
\usepackage{url}
\usepackage{hyperref}
\usepackage{CJK}
\usepackage{bm}  
\usepackage{soul}
\usepackage{textcomp}  
{}
{}
{}
\usepackage{graphicx}
\usepackage{algpseudocode}
\usepackage{amsmath}
\usepackage[normalem]{ulem} 
\usepackage{amssymb}
\usepackage{mathrsfs}
\usepackage{multirow}
\usepackage{xcolor}         
\usepackage{subfigure}
\makeatletter
\usepackage{rotating} 
\usepackage{lscape}
\newif\if@restonecol
\makeatother

\usepackage[linesnumbered,ruled,vlined]{algorithm2e}
\usepackage{algpseudocode}
\usepackage{multirow}
\usepackage{hyperref}
\hypersetup{
    colorlinks=true,
    linkcolor=blue,
    filecolor=magenta,      
    urlcolor=cyan,
    pdftitle={Overleaf Example},
    pdfpagemode=FullScreen,
    }
\usepackage{cite}
\usepackage{booktabs}
\ifCLASSINFOpdf
\else
\fi

\begin{document}

\title{Learning Spatially Adaptive Structural Coordination for Underwater Salient Object Detection}
\author{Lin Hong,~\IEEEmembership{Member,~IEEE,}
 Chenhui Wang, Linan Deng, Yuning Cui, Yu Zhang, Xin Wang,~\IEEEmembership{Member,~IEEE,} \\
   Bojian Zhang, Xingchen Yang, and Fumin Zhang,~\IEEEmembership{Fellow,~IEEE} 
\thanks{$\bullet$ Lin Hong, Linan Deng, and Fumin Zhang are with the Department of Electronic and Computer Engineering, The Hong Kong University of Science and Technology, Hong Kong, China.}
\thanks{$\bullet$ Chenhui Wang and Xin Wang are with the School of Robotics and Advanced Manufacture, Harbin Institute of Technology, Shenzhen, China. }
\thanks{$\bullet$ Yuning Cui and Yu Zhang are with the School of Computation, Information
and Technology, Technical University of Munich, Munich, Germany.} 
\thanks{$\bullet$ Bojian Zhang is with the College of Computer Science \& Visual Computing and Intelligent Perception Lab, Nankai University, Tianjin, China.} 
\thanks{$\bullet$ Xingchen Yang is with the School of
Automation, Southeast University, Nanjing, China.} 

\thanks{Manuscript received xxx, 2026; revised xxx, 20xx.}}

\markboth{IJCV submission}%
{Hong \MakeLowercase{\textit{et al.}}: Learning Spatially Adaptive Structural Coordination for Underwater Salient Object Detection}

\maketitle

\begin{abstract}
Underwater salient object detection (USOD) has attracted increasing attention for underwater scene understanding and vision-guided robotic applications. However, the spatially non-uniform degradation in underwater images causes spatially varying reliability of structural cues: boundary-sensitive responses can enhance object contours but are vulnerable to degradation-induced noises, whereas region-coherent responses improve semantic completeness but may blur object boundaries. 
Existing methods rarely explicitly consider the spatial variation in structural cue
reliability under underwater image degradation.
To address this problem, this work proposes SASC-USOD, a novel framework for learning spatially adaptive structural coordination in USOD.
The proposed framework constructs two complementary structural representations with different characteristics. A boundary-sensitive representation is obtained by combining fixed Laplacian filtering with a learnable local-detail transformation to enhance discriminative boundary information, while a region-coherent representation is generated through dual-range anisotropic large-kernel contextual aggregation to capture long-range structural consistency.
A spatial coordination module is then introduced to estimate the relative reliability of these structural representations and adaptively coordinate their contributions according to image content. 
Extensive experiments on the USOD10K and USOD benchmarks demonstrate
that SASC-USOD consistently outperforms existing methods, reducing MAE by 4.07\% and 23.53\% compared with the strongest competing method, respectively.
Moreover, its lightweight variant runs at 21 FPS on an NVIDIA Jetson
TX2 NX, demonstrating its capability for onboard underwater robotic perception.
The code is available at~\url{https://github.com/LinHong-HIT/SASC-USOD}.
\end{abstract}
\IEEEpeerreviewmaketitle

\section{Introduction}
\IEEEPARstart{U}{n}derwater salient object detection (USOD) aims to localize and segment visually distinctive objects in complex underwater environments~\cite{730558,cheng2014global}. As an important visual perception task for underwater scene
understanding, USOD can support various vision-guided robotic
applications, including marine environmental monitoring~\cite{khan2023deep}, underwater benthic surveys~\cite{johnson2010saliency}, and underwater infrastructure inspection~\cite{10324358,hong2025robust}. 
However, underwater images are frequently degraded by wavelength-dependent light absorption, scattering, and spatially
non-uniform illumination, resulting in color distortion, low
contrast, and increased noise~\cite{akkaynak2019sea,hong2021wsuie}. 
These degradation factors not only weaken object boundaries and disrupt region-level appearance consistency, but also introduce misleading structural responses, making reliable salient object detection particularly challenging in underwater scenarios.

Recent deep learning-based USOD methods formulate the task as an end-to-end dense prediction problem and employ encoder--decoder architectures to generate pixel-wise saliency maps from underwater images~\cite{jin2024underwater}. 
Benefiting from the development of CNN- and Transformer-based architectures~\cite{lecun1998gradient,dosovitskiy2021image}, substantial progress has been achieved through various strategies, including underwater image enhancement~\cite{wu2024effiseanet}, multi-scale feature fusion~\cite{huang2023fusod}, and cross-level feature interaction~\cite{wu20253}. 
Nevertheless, RGB-based USOD remains challenging because both object-level semantic regions and fine-grained structural details must be inferred from degraded visual inputs alone.
Although auxiliary depth cues (i.e., object-to-camera distance) can provide complementary geometric information, acquiring reliable underwater depth measurements remains difficult, while estimated or pseudo-depth cues may introduce additional noise and error propagation~\cite{yu2023udepth,10102831}. 
Moreover, incorporating depth information increases computational overhead and system complexity, limiting the practicality of USOD models in resource-constrained underwater robotic platforms~\cite{Ji_2021_DCF,ji2020collaborative}. 
These limitations motivate the development of RGB-based methods that can exploit the structural information available in single underwater images.

To improve visual saliency prediction from a single RGB image, a common strategy is to exploit structural cues, particularly object boundaries. 
In terrestrial salient object detection, boundary information has been extensively investigated to enhance object localization and structural accuracy. 
Representative methods utilize boundary cues to refine coarse saliency predictions~\cite{qin2019basnet,tu2020edge}, facilitate interactions between boundary and salient-region features~\cite{9008371}, or explicitly model object boundaries and interior regions through separate branches~\cite{su2019selectivity,wei2020label}. 
These approaches demonstrate that boundary information provides valuable structural guidance for preserving fine details and improving object delineation.
Boundary-aware modeling has also been explored in USOD. 
TC-USOD~\cite{10102831} jointly predicted saliency masks and object boundaries through multi-task learning, while CSUNet~\cite{wei2024csunet} introduced a contour refinement module supervised by fine- and coarse-scale contour maps. 
Although these methods verify the importance of boundary cues in USOD, they mainly focus on enhancing boundary representation through auxiliary supervision or refinement strategies. 
How to effectively coordinate local boundary-sensitive information with region-level structural coherence under severe underwater degradation remains insufficiently explored.

Beyond the general challenge of jointly preserving boundary details and salient-region completeness, underwater scenes introduce a more fundamental issue: the reliability of different structural cues is spatially variable under non-uniform image degradation. 
Local structural enhancement can recover weak object boundaries; however, underwater image degradation may also generate non-semantic local variations that resemble object contours. 
Consequently, boundary-sensitive modeling may simultaneously strengthen informative boundaries and degradation-induced false responses. 
In contrast, large-range contextual aggregation can improve regional continuity and semantic completeness by exploiting broader contextual dependencies. 
Nevertheless, reduced foreground--background contrast caused by underwater image degradation may make neighboring regions across true object boundaries exhibit similar appearances, causing excessive contextual aggregation to propagate foreground responses into background areas and blur object transitions. 
Therefore, boundary-sensitive and region-coherent structural cues exhibit complementary yet spatially varying reliability in underwater images, making fixed or spatially uniform structural fusion insufficient for USOD.

To address this challenge, this paper proposes SASC-USOD, a novel RGB-based USOD framework based on spatially adaptive structural coordination. 
The key idea is to learn where and when different structural cues should be emphasized under spatially varying underwater image degradation. 
Specifically, SASC-USOD adopts a dual-path structural modeling strategy to construct complementary boundary-sensitive and region-coherent representations from a single underwater image. 
A spatial coordination module is introduced to estimate the spatial reliability of these structural representations and adaptively coordinate their contributions according to local image characteristics. 
Furthermore, cooperative structural supervision is designed to preserve the complementary behaviors of the two structural representations and regularize the coordination process. 
By explicitly modeling the spatial variation of structural cue reliability, SASC-USOD effectively exploits complementary structural information for robust salient object detection in underwater environments. The main contributions of this work are summarized as follows:

\begin{itemize}
\item This paper proposes SASC-USOD, an RGB-based USOD framework based on spatially adaptive structural coordination. It employs a dual-path structural modeling strategy to construct complementary boundary-sensitive and region-coherent representations from underwater RGB images, and learns a spatially adaptive coordination mechanism to regulate their relative contributions according to image content.

\item Extensive comparative experiments demonstrate that the proposed SASC-USOD achieves state-of-the-art performance on the USOD10K and USOD benchmarks, and ablation studies demonstrate the effectiveness of the proposed design within the SASC-USOD framework. Moreover, it runs at 21 FPS on an NVIDIA Jetson TX2 NX, demonstrating its capability for onboard underwater robotic perception.
\end{itemize}

The rest of this paper is organized as follows: Section~\ref{sec:rw} reviews related work. Section~\ref{sec:method} presents the proposed SASC-USOD method. Section~\ref{Sec:exp} provides the experimental settings, quantitative and qualitative results, ablation studies, and real-world onboard applications. Section~\ref{sec:con} concludes this paper.

\section{Related Works}
\label{sec:rw}
\subsection{Underwater Salient Object Detection}
\label{usod}
Recent deep learning-based USOD methods commonly adopt encoder--decoder architectures based on CNNs, Transformers, or hybrid frameworks to learn hierarchical representations and generate pixel-wise saliency masks~\cite{10102831,jin2024underwater}. 
To improve RGB-based USOD, existing studies have explored various strategies from the perspectives of underwater image degradation handling and feature representation learning. 
For example, physics-inspired representation learning has been introduced to alleviate wavelength-dependent color attenuation in underwater image~\cite{liu2025hdanet}, while turbidity--similarity decoupling and feature-consistent mutual learning have been proposed to enhance robustness under degraded visibility~\cite{11339371}. 
Other representative methods exploit blurriness-aware representations~\cite{peng2024blurriness}, boundary-aware learning~\cite{wei2024csunet}, attention-based multi-scale interaction~\cite{wu20253}, or heterogeneous multi-branch feature modeling~\cite{zha2025heterogeneous} to improve salient-object localization and segmentation.
Another research direction introduces depth information as complementary geometric guidance for USOD. 
IF-USOD~\cite{yuan2025if} integrated RGB and depth features through cross-modal interaction and cross-scale collaborative learning, while UDF-Net~\cite{liu2025detecting} combined monocular depth estimation with attention-based RGB--depth fusion and task-driven joint optimization. 
Depth cues have also been explored through self-paced learning and depth-emphasis strategies~\cite{jin2024underwater}. 
Although depth information provides additional geometric constraints for separating foreground objects from complex backgrounds, acquiring reliable underwater depth remains challenging, and large-scale USOD datasets with accurate depth annotations are still limited. 
Consequently, RGB-D-based methods often rely on estimated or pseudo-depth cues~\cite{10102831}, which may introduce additional noise and error propagation.
Therefore, effectively exploiting structural and semantic information from a single underwater RGB image remains crucial for robust USOD.

\subsection{Boundary-aware Visual Saliency Modeling}
Boundary-aware modeling has been widely investigated in salient object detection due to the importance of accurate object delineation and fine structural preservation. 
One representative direction incorporates explicit boundary cues into saliency mask generation and refinement. 
BASNet~\cite{qin2019basnet} employed a predict--refine architecture with a hybrid loss to progressively improve region completeness and boundary quality, while edge-aware refinement methods utilized boundary information to recover fine structural details~\cite{tu2020edge}. 
Another direction focuses on enhancing the interaction between boundary and saliency representations. 
EGNet~\cite{9008371} combined local edge cues with global location information and integrates them with multi-scale saliency features for salient object detection. 
Recent studies further strengthen the interaction between edge and semantic representations through dual semantic interaction, edge refinement, and edge-guided cross-attention mechanisms~\cite{dsinet2025,secanet2026}. 
Beyond using boundaries as auxiliary guidance, several methods explicitly distinguish boundary-related and region-related representations. 
\cite{su2019selectivity} employed separate streams to capture selective boundary responses and invariant interior representations with transition-region compensation. 
\cite{wei2020label} decoupled saliency labels into body and detail maps to separately supervise region-oriented and detail-oriented predictions, which are subsequently integrated for saliency mask generation. 
These studies indicate that boundary structures and interior regions possess complementary characteristics and may benefit from differentiated modeling rather than homogeneous feature processing.
Boundary-aware structural modeling has also been explored in USOD. 
TC-USOD~\cite{10102831} jointly learned underwater visual saliency prediction and boundary detection through multi-task supervision to improve salient object localization and boundary delineation, whereas CSUNet~\cite{wei2024csunet} introduced a dedicated contour refinement module supervised by fine- and coarse-scale contour maps to enhance structural details. 
These methods demonstrate the effectiveness of boundary guidance in enhancing structural perception for USOD under degraded underwater conditions. Despite these advances, existing methods mainly exploit structural information through auxiliary supervision, separate prediction branches, feature interaction, or refinement strategies. 
However, how to coordinate boundary-sensitive information and region-level structural coherence under spatially varying underwater image degradation remains insufficiently investigated.

\section{SASC-USOD Method}
\label{sec:method}
Given an underwater RGB image 
$\mathbf{I}\in\mathbb{R}^{3\times H\times W}$ 
and its corresponding binary saliency mask 
$\mathbf{M}\in\{0,1\}^{1\times H\times W}$, 
the proposed SASC-USOD aims to learn a function:
\begin{equation}
\hat{\mathbf{M}} = \mathcal{F}_{\Theta}(\mathbf{I}),
\end{equation}
where $\mathcal{F}_{\Theta}(\cdot)$ denotes the proposed SASC-USOD model parameterized by $\Theta$, and 
$\hat{\mathbf{M}}\in[0,1]^{1\times H\times W}$ represents the predicted saliency map.

\begin{figure*}[ht]
\centering
\includegraphics[width=0.99\linewidth]{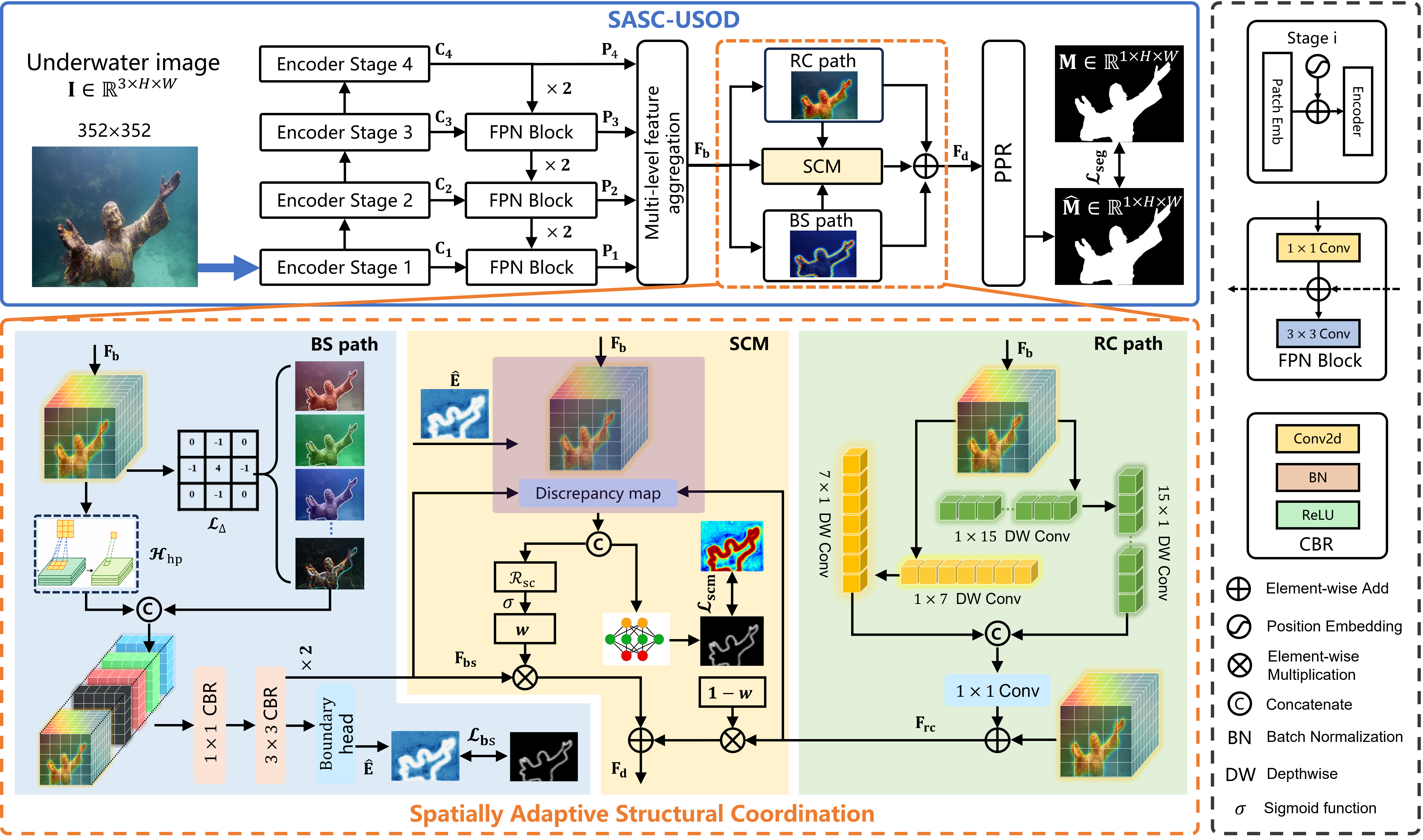}
\caption{
Overall architecture of the proposed SASC-USOD.
Given an underwater RGB image $\mathbf{I}$, the encoder first extracts
hierarchical features and produces a shared structural representation
$\mathbf{F}_{\mathrm{b}}$.
A dual-path structural modeling strategy then derives complementary
boundary-sensitive (BS) and region-coherent (RC) representations.
The spatial coordination module (SCM) adaptively coordinates these two
structural representations according to local image characteristics.
The progressive prediction refinement (PPR) module gradually recovers
the saliency mask, while cooperative structural supervision
facilitates complementary structural representation learning and stable the SCM
during training.
}
\label{fig:framework}
\end{figure*}

As illustrated in Fig.~\ref{fig:framework}, SASC-USOD adopts an encoder--decoder architecture built upon spatially adaptive structural coordination. 
The framework consists of three key components:
1) a dual-path structural modeling strategy that constructs complementary boundary-sensitive (BS) and region-coherent (RC) representations;
2) a spatial coordination module (SCM) that learns the spatial reliability of different structural representations and adaptively coordinates their contributions;
and 3) a progressive prediction refinement (PPR) module that progressively reconstructs saliency maps.
Furthermore, cooperative structural supervision is introduced to encourage the two complementary path behaviors and stabilize their coordination process.

\subsection{Shard Hierarchical Feature Extraction and Aggregation}
SASC-USOD first extracts hierarchical visual features from the input underwater RGB image using a pretrained encoder:
\begin{equation}
\left\{
\mathbf{C}_{1},
\mathbf{C}_{2},
\mathbf{C}_{3},
\mathbf{C}_{4}
\right\}
=
\mathcal{E}(\mathbf{I}),
\end{equation}
where $\mathcal{E}(\cdot)$ is the encoder network and 
$\mathbf{C}_{i}$ is the feature map extracted from the $i$-th encoder stage.

Since structural cues at different spatial scales provide complementary information for underwater visual saliency prediction, multi-scale encoder features are aggregated to construct a shared representation for structural modeling. Specifically, the feature maps from different stages are first projected into a unified channel dimension:
\begin{equation}
\tilde{\mathbf{C}}_{i}
=
\phi_{i}(\mathbf{C}_{i}),
\qquad
i\in\{1,2,3,4\},
\end{equation}
where $\phi_i(\cdot)$ represents a $1\times1$ convolution.

A top-down feature pyramid is then constructed to progressively incorporate high-level semantic information and low-level spatial details:
\begin{align}
\mathbf{P}_{4} &= \tilde{\mathbf{C}}_{4}, \\
\mathbf{P}_{3} &= \tilde{\mathbf{C}}_{3} + \mathcal{U}(\mathbf{P}_{4}), \\
\mathbf{P}_{2} &= \tilde{\mathbf{C}}_{2} + \mathcal{U}(\mathbf{P}_{3}), \\
\mathbf{P}_{1} &= \tilde{\mathbf{C}}_{1} + \mathcal{U}(\mathbf{P}_{2}),
\end{align}
where $\mathcal{U}(\cdot)$ denotes bilinear upsampling.

Finally, the pyramid features are resized to the spatial resolution of $\mathbf{P}_{1}$ and concatenated along the channel dimension to obtain the shared base representation:
\begin{equation}
\mathbf{F}_{\mathrm{b}}
=
\Phi_{\mathrm{fuse}}
\left(
[
\mathbf{P}_{1},
\mathcal{U}_{2}(\mathbf{P}_{2}),
\mathcal{U}_{3}(\mathbf{P}_{3}),
\mathcal{U}_{4}(\mathbf{P}_{4})
]
\right),
\end{equation}
where $\mathbf{F}_{\mathrm{b}}$ denotes shared base representation, $[\cdot]$ is channel-wise concatenation, 
$\mathcal{U}_{i}(\cdot)$ resizes $\mathbf{P}_{i}$ to the resolution of $\mathbf{P}_{1}$, 
and $\Phi_{\mathrm{fuse}}(\cdot)$ consists of a $1\times1$ convolution followed by GroupNorm and ReLU.


\subsection{Spatially Adaptive Structural Coordination}
\label{dss}

Given the shared representation $\mathbf{F}_{\mathrm{b}}$, SASC-USOD aims to
model two complementary structural cues: BS representation for preserving
local object boundaries and RC representation for maintaining semantic
consistency within salient regions.
However, the reliability of these structural cues is spatially variable in
underwater images with non-uniform degradations.
Therefore, SASC-USOD introduces a spatial coordination module to adaptively
coordinate the two complementary structural cues
according to local image content.

\subsubsection{BS Representation}

To construct the BS representation, SASC-USOD first extracts local
structural discontinuities associated with object boundaries.
Although high-frequency responses are beneficial for recovering weak
object boundaries, underwater image degradation may also introduce
non-semantic local variations with similar responses.
Consequently, relying solely on a fixed high-pass operator may produce
unstable boundary representations across different underwater scenes.
To achieve both structural sensitivity and local adaptability, a fixed
channel-wise Laplacian operator is combined with a learnable local-detail
transformation:
\begin{equation}
\mathbf{E}_{\mathrm{h}}
=
\mathcal{L}_{\Delta}(\mathbf{F}_{\mathrm{b}})
+
\mathcal{H}_{\mathrm{ld}}(\mathbf{F}_{\mathrm{b}}),
\end{equation}
where $\mathcal{L}_{\Delta}(\cdot)$ is a fixed channel-wise
Laplacian operator, and $\mathcal{H}_{\mathrm{ld}}(\cdot)$ is the
learnable local-detail transformation, which consists of a
depthwise $3\times3$ convolution and a pointwise $1\times1$
convolution, where each convolution is followed by batch normalization
and ReLU.

The extracted local structural responses are then integrated with the
$\mathbf{F}_{\mathrm{b}}$ to obtain the BS representation:

\begin{equation}
\mathbf{F}_{\mathrm{bs}}
=
\mathcal{B}
\left(
[
\mathbf{F}_{\mathrm{b}},
\mathbf{E}_{\mathrm{h}}
]
\right),
\end{equation}
where $[\cdot,\cdot]$ denotes channel-wise concatenation and
$\mathcal{B}(\cdot)$ consists of a $1\times1$ convolutional block
followed by two $3\times3$ convolutional blocks.
Each convolutional layer is followed by batch normalization and ReLU.

To provide structural supervision for the BS
representation, a boundary prediction head is applied to
$\mathbf{F}_{\mathrm{bs}}$:
\begin{equation}
\mathbf{A}_{\mathrm{bs}}
=
\mathcal{H}_{\mathrm{bs}}(\mathbf{F}_{\mathrm{bs}}),
\end{equation}
where $\mathcal{H}_{\mathrm{bs}}(\cdot)$ denotes the boundary prediction
head and $\mathbf{A}_{\mathrm{bs}}$ represents the boundary logit map.

The corresponding boundary probability map is obtained by:
\begin{equation}
\hat{\mathbf{E}}
=
\sigma(\mathbf{A}_{\mathrm{bs}}),
\end{equation}
where $\hat{\mathbf{E}}$ is the predicted boundary probability map.

\subsubsection{RC Representation}
To construct the RC representation, SASC-USOD introduces a dual-range
anisotropic large-kernel module to model region-level structural
coherence over different spatial ranges.
Different from the BS representation that emphasizes
local structural transitions, the RC representation aims to
capture long-range dependencies and maintain the continuity and
completeness of salient regions in degraded underwater images.

The dual-range anisotropic large-kernel module is built upon a factorized anisotropic depthwise operator.
For a kernel size $k$, the operator is defined as:
\begin{equation}
\mathcal{A}_{k}(\mathbf{X})
=
\delta
\left(
\mathrm{BN}
\left(
\mathrm{DWConv}_{k\times1}
\left(
\mathrm{DWConv}_{1\times k}(\mathbf{X})
\right)
\right)
\right),
\label{eq:anisotropic_operator}
\end{equation}
where $\mathrm{DWConv}_{1\times k}(\cdot)$ and
$\mathrm{DWConv}_{k\times1}(\cdot)$ denote horizontal and vertical
depthwise convolutions, respectively, $\mathrm{BN}(\cdot)$ denotes
batch normalization, and $\delta(\cdot)$ denotes ReLU activation.

By factorizing a two-dimensional kernel into horizontal and vertical
depthwise convolutions, $\mathcal{A}_{k}(\cdot)$ achieves an effective
$k\times k$ receptive field with reduced computational cost while
preserving directional spatial interactions.

The anisotropic operator is instantiated with kernel sizes of $7$ and
$15$ to capture contextual dependencies over different spatial ranges.
The RC representation $\mathbf{F}_{\mathrm{rc}}$ is obtained as:
\begin{equation}
\mathbf{F}_{\mathrm{rc}}
=
\mathbf{F}_{\mathrm{b}}
+
\mathcal{P}_{1\times1}
\left(
\mathcal{T}_{1\times1}
\left(
[
\mathcal{A}_{7}(\mathbf{F}_{\mathrm{b}}),
\mathcal{A}_{15}(\mathbf{F}_{\mathrm{b}})
]
\right)
\right),
\end{equation}
where $[\cdot,\cdot]$ is channel-wise concatenation,
$\mathcal{T}_{1\times1}(\cdot)$ is a $1\times1$ convolutional
block that integrates the dual-range responses and reduces their
channel dimension, and $\mathcal{P}_{1\times1}(\cdot)$ denotes a
subsequent $1\times1$ convolutional block that further transforms the
integrated features.

The residual connection preserves the original structural information
while incorporating multi-range contextual dependencies, enabling
region-level coherence enhancement without losing fine spatial details.

\subsubsection{Spatial Coordination Module}
The $\mathbf{F}_{\mathrm{bs}}$ and the $\mathbf{F}_{\mathrm{rc}}$ capture
complementary structural characteristics.
Specifically, $\mathbf{F}_{\mathrm{bs}}$ focuses on local structural
transitions and object boundaries, whereas $\mathbf{F}_{\mathrm{rc}}$
captures broader contextual dependencies to preserve regional continuity
and completeness.
However, underwater image degradation causes the reliability of these two
structural cues to vary across spatial locations.
Local appearance variations may generate boundary-like responses in
non-boundary regions, reducing the reliability of boundary-sensitive
features.
Meanwhile, reduced foreground--background contrast may cause excessive
contextual aggregation to propagate responses across actual object
boundaries, weakening region-level localization.
Therefore, a spatially adaptive coordination mechanism is required to
balance the contributions of the two structural representations according
to local image content.
To characterize the structural inconsistency between the two
representations, a channel-averaged absolute response discrepancy map
$\mathbf{D}$ is first computed:
\begin{equation}
\mathbf{D}
=
\frac{1}{C}
\sum_{c=1}^{C}
\left|
\mathbf{F}_{\mathrm{bs}}^{(c)}
-
\mathbf{F}_{\mathrm{rc}}^{(c)}
\right|,
\label{eq:branch_discrepancy}
\end{equation}
where $C$ denotes the number of feature channels.
A larger value in $\mathbf{D}$ indicates stronger structural disagreement
between the boundary-sensitive and region-coherent representations at the
corresponding spatial location.
Importantly, $\mathbf{D}$ does not explicitly determine the preferred
representation, but provides complementary evidence for learning the
spatial coordination process.

The spatial coordination map is estimated by jointly considering the
shared representation $\mathbf{F}_{\mathrm{b}}$, the predicted boundary
probability map $\hat{\mathbf{E}}$, and the structural discrepancy map
$\mathbf{D}$:
\begin{equation}
\mathbf{w}
=
\sigma
\left(
\mathcal{R}_{\mathrm{sc}}
\left(
[
\mathbf{F}_{\mathrm{b}},
\hat{\mathbf{E}},
\mathbf{D}
]
\right)
\right),
\label{eq:coordination_weight}
\end{equation}
where $\mathcal{R}_{\mathrm{sc}}(\cdot)$ denotes a lightweight
convolutional mapping consisting of a $1\times1$ convolution, a
$3\times3$ convolution, and a final $1\times1$ prediction layer.
$\sigma(\cdot)$ represents the sigmoid activation function.

The spatial coordination map
$\mathbf{w}\in[0,1]^{1\times H\times W}$ represents the spatial preference
for BS representation.
Accordingly, $1-\mathbf{w}$ represents the complementary preference for
RC representation.
The final coordinated structural representation is obtained as:
\begin{equation}
\mathbf{F}_{\mathrm{d}}
=
\mathbf{w}
\odot
\mathbf{F}_{\mathrm{bs}}
+
(1-\mathbf{w})
\odot
\mathbf{F}_{\mathrm{rc}},
\label{eq:structural_coordination}
\end{equation}
where $\odot$ denotes element-wise multiplication.

\subsection{Progressive Prediction Refinement}
Given the coordinated structural representation
$\mathbf{F}_{\mathrm{d}}$, SASC-USOD decodes the saliency
map from coarse semantic localization to fine spatial refinement.
This coarse-to-fine decoding strategy aims to recover accurate salient
regions while preserving structural details provided by the preceding
structural coordination process.

First, a low-resolution saliency logit map is generated from the
$\mathbf{F}_{\mathrm{d}}$:
\begin{equation}
\mathbf{S}_{\mathrm{s}}
=
\mathcal{H}_{\mathrm{s}}(\mathbf{F}_{\mathrm{d}}),
\end{equation}
where $\mathcal{H}_{\mathrm{s}}(\cdot)$ denotes a lightweight saliency
prediction head, and $\mathbf{S}_{\mathrm{s}}$ represents the initial
low-resolution saliency logit map.

The initial prediction is then combined with the $\mathbf{F}_{\mathrm{d}}$ to generate a coarse full-resolution map:
\begin{equation}
\mathbf{S}_{\mathrm{c}}
=
\mathcal{H}_{\mathrm{c}}
\left(
[
\mathbf{F}_{\mathrm{d}},
\mathbf{S}_{\mathrm{s}}
]
\right),
\end{equation}
where $\mathcal{H}_{\mathrm{c}}(\cdot)$ is the coarse saliency
prediction head and $\mathbf{S}_{\mathrm{c}}$ is the coarse
full-resolution saliency logit map.

To further recover fine spatial details, a residual refinement strategy
is introduced by incorporating the RGB image and low-level features:
\begin{equation}
\mathbf{S}_{\mathrm{f}}
=
\mathbf{S}_{\mathrm{c}}
+
\mathcal{H}_{\mathrm{r}}
\left(
[
\mathbf{I},
\mathbf{S}_{\mathrm{c}},
\rho_{\mathrm{f}}(\mathbf{P}_{1})
]
\right),
\end{equation}
where $\rho_{\mathrm{f}}(\cdot)$ performs channel reduction and
full-resolution resizing on $\mathbf{P}_{1}$, and
$\mathcal{H}_{\mathrm{r}}(\cdot)$ denotes the residual refinement head
that predicts spatial corrections for $\mathbf{S}_{\mathrm{c}}$.

Finally, the predicted saliency probability map is obtained as:
\begin{equation}
\hat{\mathbf{M}}
=
\sigma(\mathbf{S}_{\mathrm{f}}),
\end{equation}

\subsection{Cooperative Structural Supervision}
The cooperative structural supervision strategy is designed to promote complementary learning between the BS and RC representations while stabilizing the spatial coordination process.
It jointly supervises saliency map prediction, boundary-sensitive structural
learning, and coordination map estimation during training.

The PPR module produces three saliency logit maps:
the final full-resolution logits $\mathbf{S}_{\mathrm{f}}$, the coarse
full-resolution logits $\mathbf{S}_{\mathrm{c}}$, and the
low-resolution logits $\mathbf{S}_{\mathrm{s}}$.
Let $\mathbf{M}_{\mathrm{s}}$ denote the ground-truth mask
$\mathbf{M}$ resized to the spatial resolution of
$\mathbf{S}_{\mathrm{s}}$.
The saliency map prediction loss is defined as:
\begin{equation}
\begin{aligned}
\mathcal{L}_{\mathrm{seg}}
=&
\lambda_{\mathrm{f}}
\mathcal{L}_{\mathrm{str}}
\left(
\mathbf{S}_{\mathrm{f}},\mathbf{M}
\right)
+
\lambda_{\mathrm{c}}
\mathcal{L}_{\mathrm{str}}
\left(
\mathbf{S}_{\mathrm{c}},\mathbf{M}
\right)
\\
&+
\lambda_{\mathrm{s}}
\mathcal{L}_{\mathrm{str}}
\left(
\mathbf{S}_{\mathrm{s}},\mathbf{M}_{\mathrm{s}}
\right),
\end{aligned}
\end{equation}
where $\mathcal{L}_{\mathrm{str}}(\cdot)$ denotes the 
structural loss, consisting of a binary cross-entropy loss and an IoU loss
to optimize pixel-level classification and region-level structural
consistency, respectively. $\lambda_{\mathrm{f}}$, $\lambda_{\mathrm{c}}$, and
$\lambda_{\mathrm{s}}$ control the contributions of the final, coarse,
and low-resolution saliency map predictions, respectively.

To encourage the BS representation to preserve boundary-sensitive
characteristics, a ground-truth boundary map $\mathbf{E}$ is extracted
from $\mathbf{M}$ and resized to the spatial resolution of
$\mathbf{A}_{\mathrm{bs}}$, resulting in $\mathbf{E}_{\mathrm{s}}$.
The boundary supervision loss is defined as:
\begin{equation}
\mathcal{L}_{\mathrm{bs}}
=
\mathcal{L}_{\mathrm{bce}}^{\mathrm{logit}}
\left(
\mathbf{A}_{\mathrm{bs}},
\mathbf{E}_{\mathrm{s}}
\right)
+
\mathcal{L}_{\mathrm{dice}}
\left(
\hat{\mathbf{E}},
\mathbf{E}_{\mathrm{s}}
\right),
\end{equation}
where $\mathcal{L}_{\mathrm{bce}}^{\mathrm{logit}}(\cdot)$ denotes
binary cross-entropy with logits and $\mathcal{L}_{\mathrm{dice}}(\cdot)$
denotes the Dice loss.

Furthermore, a structural prior is introduced to guide the learning of
the spatial coordination process.
Specifically, a soft coordination target is generated from the boundary
map $\mathbf{E}_{\mathrm{s}}$:
\begin{equation}
\mathbf{R}_{\mathrm{s}}^{*}
=
\mathrm{AvgPool}
\left(
\mathrm{MaxPool}
\left(
\mathbf{E}_{\mathrm{s}}
\right)
\right).
\end{equation}

The $\mathrm{MaxPool}$ operation expands the supervised boundary regions,
while $\mathrm{AvgPool}$ provides smooth transitions around these regions.
The resulting target does not enforce a deterministic selection between
the two structural representations, but provides a boundary-aware prior
for learning spatially adaptive coordination.
Since $\mathbf{w}$ controls the contribution of
$\mathbf{F}_{\mathrm{bs}}$, the spatial coordination loss is defined as:
\begin{equation}
\mathcal{L}_{\mathrm{scm}}
=
\mathcal{L}_{\mathrm{bce}}^{\mathrm{logit}}
\left(
\mathbf{A}_{\mathrm{sc}},
\mathbf{R}_{\mathrm{s}}^{*}
\right),
\end{equation}
where $\mathbf{A}_{\mathrm{sc}}$ is the coordination logits
predicted by the SCM.

The overall training objective is formulated as:
\begin{equation}
\mathcal{L}
=
\mathcal{L}_{\mathrm{seg}}
+
\lambda_{\mathrm{bs}}
\mathcal{L}_{\mathrm{bs}}
+
\lambda_{\mathrm{scm}}
\mathcal{L}_{\mathrm{scm}},
\end{equation}
where $\lambda_{\mathrm{bs}}$ and $\lambda_{\mathrm{scm}}$ represent the
weighting coefficients for $\mathcal{L}_{\mathrm{bs}}$ and
$\mathcal{L}_{\mathrm{scm}}$, respectively.

\section{Experiments}
\label{Sec:exp}

\subsection{Experimental Setup}

\subsubsection{Implementation Details}
The proposed SASC-USOD is implemented in PyTorch
with a pretrained PVTv2~\cite{wang2022pvt} adopted as the encoder
backbone.
To investigate the effectiveness under different model capacities, two
variants are considered: SASC-USOD-T equipped with PVTv2-B3 and
SASC-USOD-S equipped with PVTv2-B5.
Unless otherwise specified, SASC-USOD-S is used as the default model.
The model is trained on the USOD10K dataset~\cite{10102831}, and all
input images are resized to $352\times352$ during both training and
inference.
SASC-USOD is trained for 50 epochs on an NVIDIA RTX 4090 GPU using the
AdamW optimizer with an initial learning rate of $1\times10^{-4}$, a
weight decay of $1\times10^{-2}$, and a cosine annealing learning-rate
schedule with a minimum learning rate of $1\times10^{-6}$.
$\mathcal{L}_{\mathrm{str}}$ is
applied to the final full-resolution, coarse full-resolution, and
low-resolution predictions with weights of 4.0, 0.25, and 0.25,
respectively.
The $\mathcal{L}_{\mathrm{bs}}$ and the $\mathcal{L}_{\mathrm{scm}}$ are further introduced to
preserve boundary-sensitive characteristics and stabilize the spatial coordination process, respectively.

\subsubsection{Datasets and Evaluation Metrics}
Quantitative evaluation of the proposed SASC-USOD is conducted on two
widely used USOD benchmark datasets: USOD10K~\cite{10102831} and
USOD~\cite{islam2022svam}.
Following common evaluation protocols, five metrics are adopted:
S-measure ($S_{\mathrm{m}}$)~\cite{8237749}, maximum E-measure
($E_{\xi}^{\mathrm{max}}$)~\cite{2018Enhanced}, maximum F-measure
($\mathrm{max}F$)~\cite{5206596}, mean absolute error (MAE)~\cite{6247743},
and the precision--recall (PR) curve.
Specifically, $S_{\mathrm{m}}$ evaluates the structural similarity between
the predicted saliency map and the ground truth from both region-aware
and object-aware perspectives.
$E_{\xi}^{\mathrm{max}}$ measures both local pixel-level alignment and
global image-level consistency.
$\mathrm{max}F$ computes the harmonic mean of precision and recall over
different thresholds, reflecting the overall foreground--background
discrimination capability.
MAE measures the average absolute difference between the predicted
saliency map and the ground-truth mask.
The PR curve evaluates the precision--recall trade-off under different
thresholds, providing a comprehensive assessment of saliency prediction
robustness.

\subsubsection{Compared Methods}
SASC-USOD is compared with 40 representative state-of-the-art (SOTA)
methods for comprehensive evaluation.
The competitor set covers four categories: 
17 RGB SOD methods include BASNet~\cite{qin2019basnet}, EGNet~\cite{9008371}, LDF~\cite{wei2020label}, F3Net~\cite{F3Net}, PFPN~\cite{2020Progressive}, MINet~\cite{MINet-CVPR2020}, DASNet~\cite{zhao2020DASNet}, PFSNet~\cite{ma2021pyramidal}, MFNet~\cite{piao2021mfnet}, CTDNet~\cite{zhao2021complementary}, PSGLoss~\cite{yang2021progressive}, VST~\cite{Liu_2021_ICCV}, CSNet~\cite{21PAMI-Sal100K}, ISNet~\cite{zhu2024isnet}, ABiU-Net~\cite{qiu2024abiunet}, iGAN~\cite{mao2025igan}, and SDNet-A~\cite{su2025rapid}; 
17 RGB-D SOD methods include JL-DCF~\cite{fu2020jl}, UC-Net~\cite{Zhang2020UCNet}, S2MA~\cite{liu2020learning}, BBS-Net~\cite{fan2020bbs}, DANet~\cite{zhao2020single}, DCF~\cite{Ji_2021_DCF}, SPNet~\cite{zhou2021specificity}, HAINet~\cite{Li_2021_HAINet}, TriTransNet~\cite{Liu2021TriTransNetRS}, D3Net~\cite{fan2020rethinking}, BTS-Net~\cite{Zhang2021BTSNet}, CDINet~\cite{Zhang2021CDINet}, CIR-Net~\cite{cong2022cirnet}, HiDAnet~\cite{wu2023hidanet}, PICR-Net~\cite{cong2023picrnet}, CATNet~\cite{sun2024catnet}, and SATNet~\cite{10970446};
2 RGB USOD methods include SVAM-Net~\cite{islam2022svam} and HDANet~\cite{liu2025hdanet}; 4 RGB-D USOD methods include TC-USOD~\cite{10102831}, SPDENet~\cite{jin2024underwater}, FSCDiff~\cite{li2025fscdiff}, and SearchNet~\cite{11339371}. 
This comparison set covers diverse model paradigms, including
CNN-based, Transformer-based, and diffusion-based architectures.
To ensure a fair comparison, all competing methods are
retrained on the USOD10K training set using their official
implementations.



\begin{table}[t]
\centering
\caption{Quantitative evaluation of the SASC-USOD by comparing 40 SOTA methods. The performance of these methods is tested on the USOD10K and the USOD datasets. The top-3 results are marked in \textcolor{red}{red}, \textcolor{blue}{blue}, and \textcolor{cyan}{cyan}, respectively.}
\label{tab:Quantitative}

\resizebox{\columnwidth}{!}{%
\begin{tabular}{r|c||cccc||cccc}

\toprule
\multirow{2}{*}{\textbf{Methods}} & \multirow{2}{*}{\textbf{Pub.}}
&\multicolumn{4}{c}{\textbf{USOD10K} (1026 images)}  
&\multicolumn{4}{c}{\textbf{USOD} (300 images)}  \\
\cmidrule(r){3-6} \cmidrule(r){7-10} 

& & $S_\mathrm{m} \uparrow $ & $E_{\xi}^{max} \uparrow$ & $\text{max}F  \uparrow$ & $\text{MAE} \downarrow$ 
& $S_\mathrm{m} \uparrow $ & $E_{\xi}^{max} \uparrow$ & $\text{max}F  \uparrow$ & $\text{MAE} \downarrow$  \\

\hline
\hline

\rowcolor[gray]{0.95}\multicolumn{10}{c}{{RGB SOD methods}}\\
\hline
\rowcolor[RGB]{220,235,247}BASNet & CVPR\_19
                            &.8937  & .9378  &.8849 & .0352 
                            &.8941  & .9311  &.9067 & .0490\\   
                            
EGNet & ICCV\_19
                            &.9125  & .9488  &.9040 & .0291 
                            &.8947  & .9278  &.9077 & .0534\\
                            
                            
\rowcolor[RGB]{220,235,247}LDF & CVPR\_20
                            &.9135  & .9574  &.9173 & .0260 
                            &.9027  & .9380  &.9205 & .0447  \\  
                            
F3Net & AAAI\_20
                            &.9140  & .9599  &.9171 & .0251 
                            &.8995 & .9355  &.9155  & .0448  \\ 
                            
\rowcolor[RGB]{220,235,247}PFPN & AAAI\_20
                            &.9090  & .9547  &.9055 & .0302 
                            &.8940  & .9351  &.9077 & .0516  \\ 
                            
MINet & CVPR\_20
                            &.9105  & .9501  &.9072 & .0287
                            &.8968  & .9329  &.9120 & .0471  \\ 
                            
\rowcolor[RGB]{220,235,247}DASNet & ACMMM\_20
                            &.9204  & .9603  &.9212 &.0245 
                            &.9004  & .9350  &.9147 &.0458 \\ 

CSNet & TPAMI\_22
                            &.8595  & .9178  &.8462 & .0548
                            &.8557  & .9051  &.8594 & .0837\\  
                            
\rowcolor[RGB]{220,235,247}CTDNet & ACMMM\_21
                            &.9085  & .9531  &.9073 & .0285
                            &.8924  & .9283  &.9015 & .0476\\  
                            
MFNet & ICCV\_21
                            &.8425  & .9146  &.8193 & .0512 
                            &.8421  & .9105  &.8509 & .0759  \\ 
                            
\rowcolor[RGB]{220,235,247}PFSNet & AAAI\_21
                            &.8983  & .9421  &.8966 & .0370
                            &.8932  & .9281  &.9077 & .0511  \\  
                            
PSGLoss & TIP\_21
                            &.8640  & .9078  &.8508 & .0417
                            &.8790  & .9262  &.9038 & .0496  \\
\rowcolor[RGB]{220,235,247}VST & ICCV\_21
                            &.9136  &{.9614} &.9108 & .0267
                            &.8941  & .9322  &.9039 & .0537  \\ 
ABiU-Net & TCSVT\_24
                            &.9173  & .9628  &.9192 & .0258
                            &.8966  & .9343  &.9124 & .0502  \\

\rowcolor[RGB]{220,235,247}ISNet & TCSVT\_24
                            &.9222  & .9635  &.9203 & .0249
                            &.8995  & .9352  &.9133 & .0473  \\

iGAN & TCSVT\_25
                            &.9055  & .9596  &.9033 & .0264
                            &.8926  & .9372  &.9071 & .0465  \\
\rowcolor[RGB]{220,235,247}SDNet-A & TPAMI\_25
                            &.8882  & .9437  &.8905 & .0340
                            &.8845  & .9209  &.8928 & .0556  \\                 

\hline
\rowcolor[gray]{0.95}\multicolumn{10}{c}{{RGB-D SOD methods}}\\
\hline

JL-DCF & CVPR\_20
                            &.9062  & .9485  &.8978 & .0300 
                            &.8909  & .9292  &.9044 & .0540  \\
                            
\rowcolor[RGB]{220,235,247}UCNet & CVPR\_20
                            &.8997  & .9463  &.8968 & .0301
                            &.8970  & .9377  &.9158 & .0427  \\ 
                            
S2MA & CVPR\_20
                            &.8664  & .9208  &.8530 & .0558 
                            &.8627  & .9095  &.8714 & .0791 \\ 
                            
\rowcolor[RGB]{220,235,247}BBSNet & ECCV\_20
                            &.9061  & .9512  &.9056 & .0337 
                            &.8855  & .9233  &.8962 & .0596 \\ 
                            
DANet & ECCV\_20
                            &.9006  & .9449  &.8934 & .0279 
                            &.8822  & .9235  &.8987 & .0497  \\ 
                            
\rowcolor[RGB]{220,235,247}DCF & CVPR\_21
                            & .9116 & .9541  & .9045 & .0312 
                            & .8929 & .9307  & .9019 & .0531 \\
                            
SPNet & ICCV\_21
                            &.9075  & .9554  &.9069 & .0280 
                            &.8870  & .9254  &.9019 & .0514  \\ 
                            
\rowcolor[RGB]{220,235,247}HAINet & ICCV\_21
                            &.9123  & .9552  &.9116 & .0279 
                            &.8957  & .9281  &.9104 & .0494  \\ 
TriTransNet & ACMMM\_21
                            &.7889  & .8479  &.7501 & .0659 
                            &.8189  & .8844  &.8452 & .0840 \\  
                            
\rowcolor[RGB]{220,235,247}D3Net & TNNLS\_21
                            &.8931  & .9413  &.8807 & .0374
                            &.8804  & .9211  &.8934 & .0615 \\  
                            
BTS-Net & ICME\_21
                            &.9093  & .9542  &.9104 & .0291
                            &.8861  & .9206  &.8987 & .0558 \\  
                            
\rowcolor[RGB]{220,235,247}CDINet & ICCV\_21
                            &.7049 & .8644  &.7362 & .0904 
                            &.6209  & .8243  &.7208 & .1585  \\ 

CIR-Net & TIP\_22
                            &.9109  & .9556  &.9105 & .0307
                            &.8907  & .9263  &.9030 & .0550  \\

\rowcolor[RGB]{220,235,247}PICR-Net & TCSVT\_23
                            &.9211  & .9640  &.9208 & .0225
                            &.8956  & .9303  &.9064 & .0460  \\

HiDAnet & TIP\_23
                            &.9162  & .9603  &.9175 & .0250
                            &.9023  & .9393  &.9174 & .0441  \\

\rowcolor[RGB]{220,235,247}CATNet & TCSVT\_24
                            &.9207  & .9657  &.9237 & .0214
                            &.8954  & .9291  &.9057 & .0463  \\
                            
SATNet      & TIP\_25
                            &.9091   & .9559   &.9086  &.0271
                            &.8900   & .9279   &.9036  &.0497  \\

\hline
\rowcolor[gray]{0.95}\multicolumn{10}{c}{{RGB-D USOD methods}}\\
\hline

\rowcolor[RGB]{220,235,247}TC-USOD & TIP\_25
                            &.9215  & .9683  &.9236 & .0201                  
                            &.8940  & .9332  &.9087 & .0462\\
                            
SPDENet & TMM\_24
                            &.9233  & .9688  &.9273 & .0199               
                            &.9001  & .9338  &.9134 & .0436\\
                            
\rowcolor[RGB]{220,235,247}FSCDiff & ACMMM\_25
                            &\textcolor{cyan}{.9321}  & \textcolor{cyan}{.9712}  &\textcolor{cyan}{.9325} &\textcolor{blue}{.0172}
                            &\textcolor{cyan}{.9058}  & .9359  &.9213 &.0404  \\  
SearchNet & TIP\_26
                            &{.9263}  & {.9709}  & {.9288} &\textcolor{cyan}{.0188}
                            &.9036  & \textcolor{cyan}{.9465}  &\textcolor{cyan}{.9227} &\textcolor{cyan}{.0391}  \\  
\hline                          
\rowcolor[gray]{0.95}\multicolumn{10}{c}{{RGB-USOD methods}}\\
\hline

\rowcolor[RGB]{220,235,247}SVAM-Net & RSS\_22
                            &.7465 & .7649  &.6451 & .0915 
                            &.8993  & .9409  &.9127 & .0517 \\ 
                            
HDANet & TCSVT\_25
                            &.9232  & .9679  &{.9266} & .0202
                            &.9030  & .9377  &{.9193} & .0418  \\

\hline   
\rowcolor[gray]{0.85}\textbf{SASC-USOD-T} & -
                            &\textcolor{blue}{.9325}  & \textcolor{blue}{.9713}   &\textcolor{blue}{.9348}  &{.0194}               
                            &\textcolor{blue}{.9178}  & \textcolor{blue}{.9489}   &\textcolor{blue}{.9278}  & \textcolor{blue}{.0343} \\
                            
\rowcolor[gray]{0.85}\textbf{SASC-USOD-S} & -
                            &\textcolor{red}{.9338}  & \textcolor{red}{.9740}   &\textcolor{red}{.9364}  &\textcolor{red}{.0165}               
                            &\textcolor{red}{.9222}  & \textcolor{red}{.9547}   &\textcolor{red}{.9335}  & \textcolor{red}{.0299} \\

\bottomrule
\end{tabular}}
\end{table}

\subsection{Comparison with SOTA Methods}

\subsubsection{Quantitative Comparisons}
The proposed SASC-USOD is quantitatively compared with 40 representative
SOTA methods on the USOD10K and USOD datasets under the same
evaluation metrics. The comparison results are summarized in
Table~\ref{tab:Quantitative}.

\begin{figure}[t]
\centering
\includegraphics[width=0.99\linewidth]{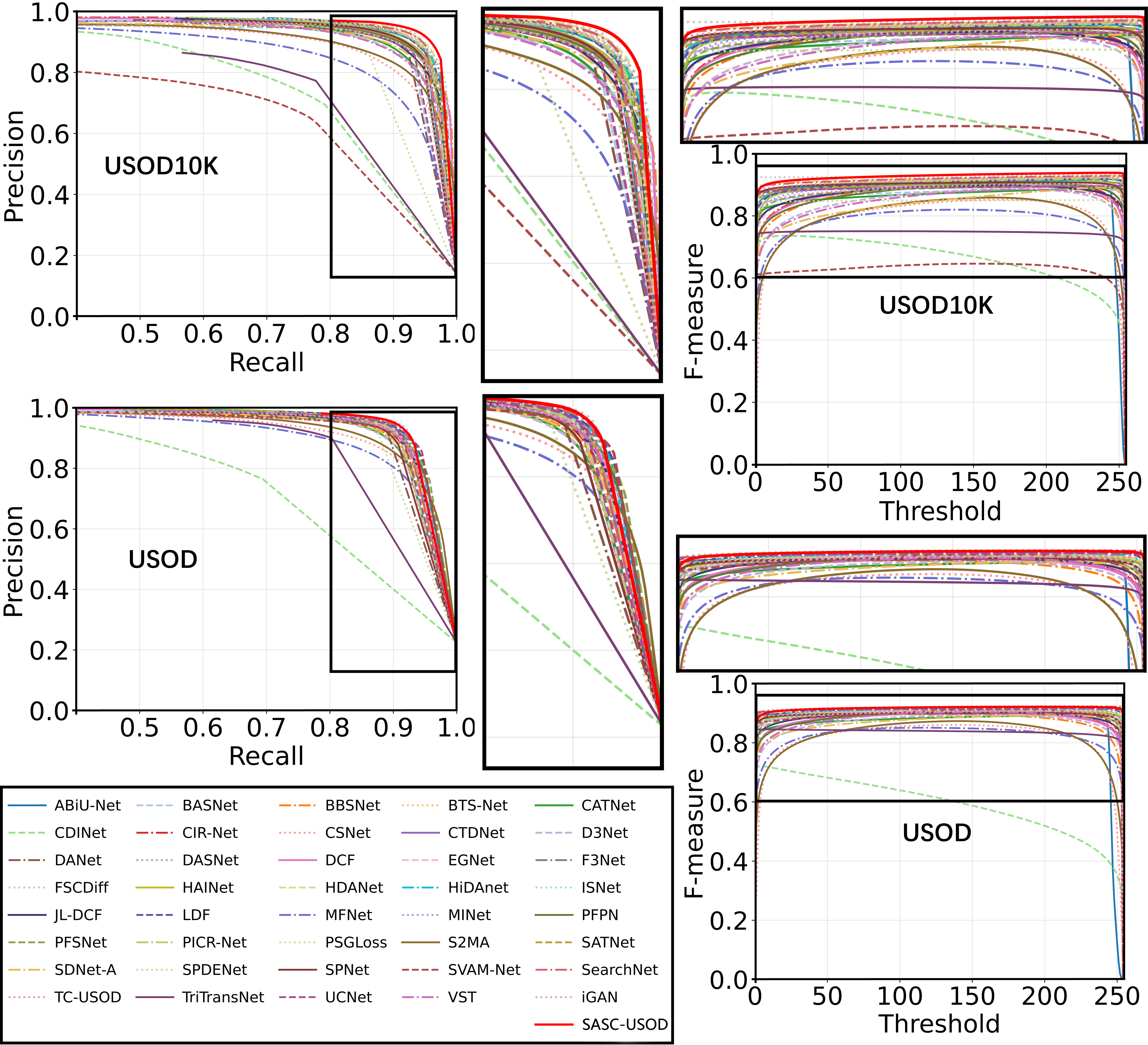}
   \caption{PR curves and F-measure curves of different methods on the USOD10K and
USOD datasets.}
\label{fig:PR_F_curve}
\end{figure}

\begin{figure*}[t]
\centering
 \includegraphics[width=1.0\linewidth]{Figures/qualitative_compare2_2_2.png} 
\caption{
Qualitative comparison of SASC-USOD with 40 SOTA methods on
USOD10K (first three rows) and USOD (last three rows).
Benefiting from spatially adaptive structural coordination, SASC-USOD
better preserves fine object boundaries while maintaining regional
coherence, producing more accurate saliency maps in challenging
underwater scenes. Zoom-in for more details. (GT: ground-truth map).
}
\label{fig:qualitative_com}
\end{figure*}

\textbf{Performance on the USOD10K dataset.}
On the USOD10K dataset, SASC-USOD achieves the best performance across
all four evaluation metrics, obtaining 0.9338 in $S_{\mathrm{m}}$,
0.9740 in $E_{\xi}^{\max}$, 0.9364 in max$F$, and 0.0165 in MAE.
Compared with FSCDiff, the strongest competing RGB-D USOD method on this
dataset, SASC-USOD improves $S_{\mathrm{m}}$ by 0.18\%,
$E_{\xi}^{\max}$ by 0.29\%, and max$F$ by 0.42\%, while reducing MAE by
4.07\%.
These consistent improvements indicate that SASC-USOD can effectively
balance boundary preservation and salient region completeness under
complex underwater environments.
Notably, SASC-USOD relies only on RGB images, while several strong
competitors, including FSCDiff and SearchNet, exploit additional depth
information.
The superior performance demonstrates that adaptive structural
coordination provides an effective alternative for exploiting intrinsic
visual cues from a single underwater image.

\textbf{Performance on the USOD dataset.}
On the USOD dataset, SASC-USOD also achieves the best performance across
all four evaluation metrics, reaching 0.9222 in $S_{\mathrm{m}}$,
0.9547 in $E_{\xi}^{\max}$, 0.9335 in max$F$, and 0.0299 in MAE.
Compared with SearchNet, the strongest competing RGB-D USOD method in
terms of $E_{\xi}^{\max}$, max$F$, and MAE, SASC-USOD improves
$E_{\xi}^{\max}$ by 0.87\% and max$F$ by 1.17\%, while reducing MAE by
23.53\%.
Compared with FSCDiff, which achieves the best competing
$S_{\mathrm{m}}$ score, SASC-USOD further improves
$S_{\mathrm{m}}$ by 1.81\%, $E_{\xi}^{\max}$ by 2.01\%, and max$F$ by
1.32\%, while reducing MAE by 25.99\%.
These results demonstrate that SASC-USOD effectively improves both
structural accuracy and pixel-level prediction quality, particularly
under challenging underwater conditions with degraded object
boundaries and ambiguous foreground--background transitions.

Fig.~\ref{fig:PR_F_curve} compares the PR and F-measure curves of
different methods on USOD10K and USOD.
SASC-USOD achieves competitive precision--recall behavior
on both datasets.
On USOD10K, SASC-USOD maintains higher precision over a broad recall
range and achieves the leading F-measure curve, which is consistent with
its highest max$F$ score in Table~\ref{tab:Quantitative}.
On USOD, SASC-USOD also maintains strong performance under different
threshold settings.
These observations further validate that the proposed spatial adaptive structural
coordination improves the robustness of saliency map prediction
rather than only optimizing performance under a specific threshold.

To further investigate the influence of encoder architectures,
SASC-USOD with different CNN- and Transformer-based backbones is evaluated,
as shown in Table~\ref{tab:backbone_comparison}.
Overall, the performance varies with backbone architectures and model
capacities, indicating that feature extraction capability influences
USOD performance.
Among CNN-based backbones, the evaluated ConvNeXt variants consistently
outperform their ResNet counterparts.
Among Transformer-based backbones, PVTv2 achieves strong overall
performance, particularly on USOD10K.
Specifically, SASC-USOD-S equipped with PVTv2-B5 achieves the best
results across all four metrics on USOD10K, while SASC-USOD-T with
PVTv2-B3 maintains competitive performance with fewer parameters.
These results demonstrate that the proposed spatially adaptive
structural coordination framework is compatible with different encoder
architectures and maintains effectiveness across different backbones.

\subsubsection{Qualitative Comparisons}
Fig.~\ref{fig:qualitative_com} presents qualitative comparisons between
SASC-USOD and 40 representative SOTA methods on the USOD10K
and USOD datasets.
Consistent with the quantitative comparisons in
Table~\ref{tab:Quantitative}, generic RGB SOD methods often struggle with
underwater image degradation.
They may generate incomplete salient regions, inaccurate object
boundaries, or false responses in complex backgrounds.
Although boundary-aware RGB SOD methods, such as BASNet and EGNet,
explicitly incorporate boundary information, their performance may still
be affected by underwater degradation, resulting in inaccurate contours
or fragmented predictions when boundary cues become unreliable.
RGB-D SOD methods benefit from additional depth information and generally
achieve improved object localization.
However, transferred depth cues may not always provide sufficient
structural guidance in underwater environments, leading to missed fine
details or coarse object boundaries in challenging cases.
Compared with generic SOD and RGB-D approaches, USOD-specific methods,
including FSCDiff, SearchNet, and HDANet, demonstrate stronger
adaptation to underwater scenarios.
Nevertheless, they may still produce fragmented salient regions, blurred
boundaries, or background false positives when underwater degradation
causes ambiguity between foreground objects and surrounding regions.
In contrast, SASC-USOD generates more complete salient regions while
preserving accurate object boundaries across diverse underwater scenes.
It also better maintains small and slender structures and suppresses
distracting background responses.
These qualitative improvements show that the proposed spatially
adaptive structural coordination effectively balances BS
and RC representation under underwater image degradation.

\begin{table}[t]
\centering
\caption{Performance and parameter comparison of SASC-USOD with different backbones on USOD10K and USOD datasets.}
\label{tab:backbone_comparison}

\resizebox{\columnwidth}{!}{%
\begin{tabular}{l|c|cccc|cccc}
\toprule
\multirow{2}{*}{Backbone} & \multirow{2}{*}{Params} 
& \multicolumn{4}{c|}{\textbf{USOD10K} (1026 images)} 
& \multicolumn{4}{c}{\textbf{USOD} (300 images)} \\
\cmidrule(lr){3-6} \cmidrule(lr){7-10}
& 
& $S_\mathrm{m} \uparrow$ 
& $E_{\xi}^{\max} \uparrow$ 
& $\text{max}F \uparrow$ 
& $\mathrm{MAE} \downarrow$
& $S_\mathrm{m} \uparrow$ 
& $E_{\xi}^{\max} \uparrow$ 
& $\text{max}F \uparrow$ 
& $\mathrm{MAE} \downarrow$ \\
\midrule
\rowcolor[RGB]{220,235,247}ResNet-50~\cite{he2016deep}          & 29.16M &.9164  & .9609 & .9173 & .0255 & .8995 & .9363 & .9073 & .0427 \\
ResNet-101~\cite{he2016deep}         & 48.15M & .9221 & .9644 & .9243 & .0235 & .9014 & .9400 & .9143 & .0418 \\
\midrule
\rowcolor[RGB]{220,235,247}ConvNeXt-Tiny~\cite{liu2022convnet}  & 32.86M &.9287  & .9685 & .9308 & .0199 &.9127  & .9448 & .9243 & .0364 \\
ConvNeXt-Small~\cite{liu2022convnet} & 54.49M & .9319 & .9707 & .9341 & .0185 & .9174 & .9487 & .9280 & .0338 \\
\rowcolor[RGB]{220,235,247}ConvNeXt-Base~\cite{liu2022convnet}  & 92.60M & .9309 & .9712 & .9341 & .0191 & .9187 & .9514 & .9313 & .0327 \\
\midrule
Swin-Tiny~\cite{liu2022swin}         & 32.83M & .9261 & .9661 & .9267 &.0218  & .9132 & .9451 & .9226 &.0350  \\
\rowcolor[RGB]{220,235,247}Swin-Small~\cite{liu2022swin}        & 54.13M & .9282 & .9679 & .9286 & .0207 & .9240 & .9572 & .9345 & .0304 \\
\midrule
PVTv2-B1~\cite{wang2022pvt}          & 17.52M & .9259 & .9664 & .9272 & .0224 & .9120 & .9458 & .9238 & .0370 \\
\rowcolor[RGB]{220,235,247}PVTv2-B3~\cite{wang2022pvt}          & 49.62M & .9325 & .9713 & .9348 & .0194 & .9178 & .9489 & .9278 &.0343  \\
\midrule
PVTv2-B5~\cite{wang2022pvt}          & 86.42M 
& \textbf{.9338} 
& \textbf{.9740} 
& \textbf{.9364} 
&\textbf{.0165} 
 &\textbf{.9222}  & \textbf{.9547}   &\textbf{.9335}  & \textbf{.0299} \\


\bottomrule
\end{tabular}%
}
\end{table}

\begin{figure}[t]
\centering
\includegraphics[width=1\linewidth]{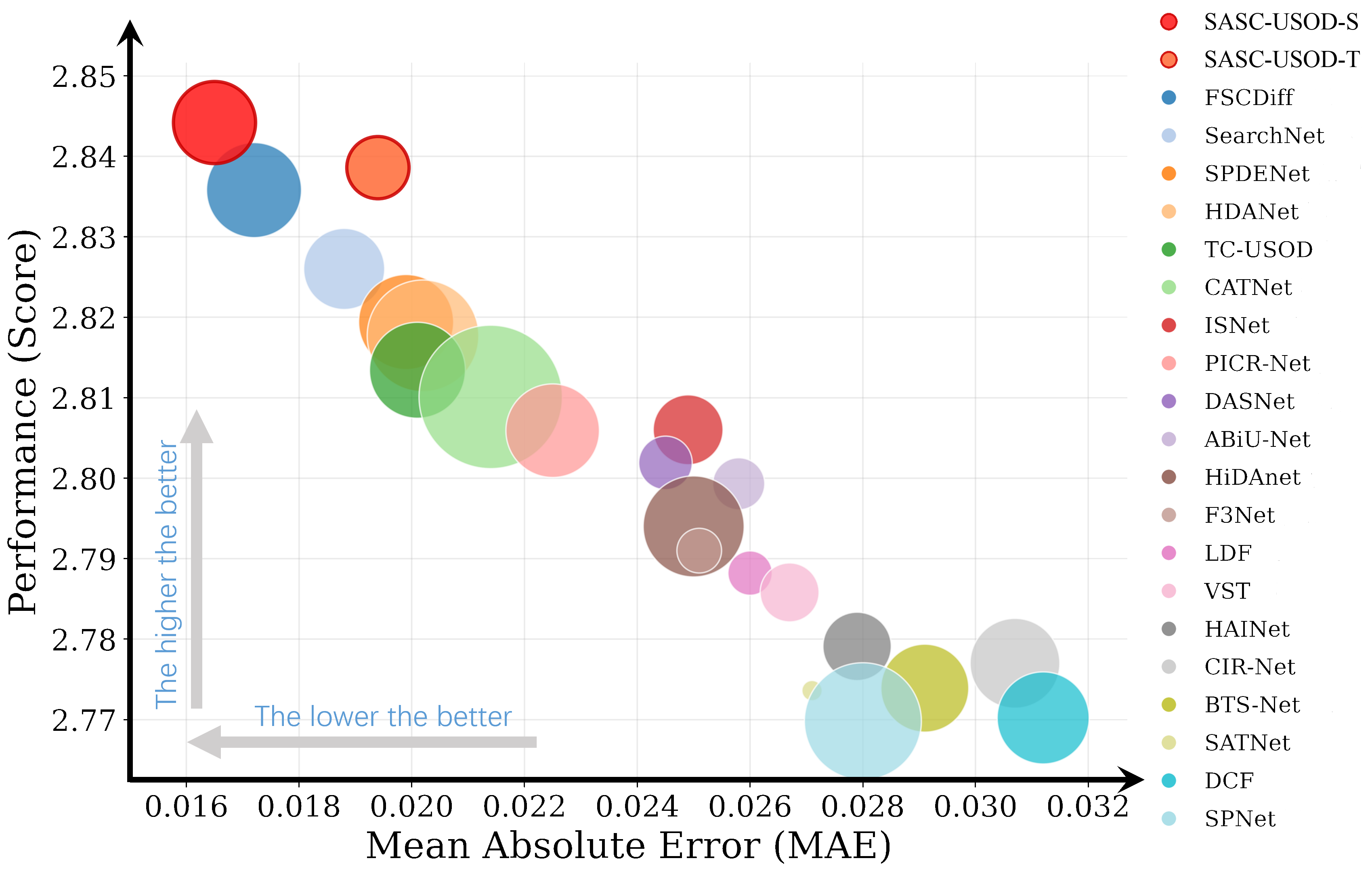}
\caption{
Visualization of model complexity and USOD performance. 
The horizontal axis denotes MAE, and the vertical axis is the sum of three positive-oriented metrics, including $S_\mathrm{m}$, $E_{\xi}^{\max}$, and max$F$. 
The circle size represents model size. 
Methods closer to the upper-left region with smaller circles indicate a better trade-off between model complexity and performance.
}
\label{fig:MCA}
\end{figure}

\subsubsection{Model Complexity and Performance Analysis}
To evaluate the deployment potential of SASC-USOD, the
relationship between model complexity and USOD performance is analyzed.
Model complexity is measured by the number of parameters.
SASC-USOD is compared with 20 representative methods on USOD10K in terms
of parameter count and quantitative performance, as shown in
Fig.~\ref{fig:MCA}.
In the visualization, MAE is plotted on the horizontal axis, while the
sum of three higher-is-better metrics, namely $S_\mathrm{m}$,
$E_{\xi}^{\max}$, and max$F$, is plotted on the vertical axis.
The circle size represents the number of model parameters.
Therefore, methods located closer to the upper-left region with smaller
circles achieve a better balance between prediction performance and
model complexity.
Compared with RGB-D USOD methods, SASC-USOD relies solely on RGB images
without requiring auxiliary depth cues or additional modality-specific
processing.
Meanwhile, compared with existing RGB-based USOD methods, SASC-USOD
achieves competitive performance with moderate model complexity.
These results demonstrate the practical deployment potential of
SASC-USOD on resource-constrained underwater robotic platforms.

\begin{table*}[t]
\centering
\caption{Generalization evaluation results of SASC-USOD on RGB SOD datasets: DUTS, DUT-OMRON, HKU-IS, ECSSD, SOD, and PASCAL-S. The top three results are highlighted in \textcolor{red}{red}, \textcolor{blue}{blue}, and \textcolor{cyan}{cyan}, respectively. ``--'' indicates that the corresponding result was not reported in the original work.}
\label{tab:USOD2SOD}
\resizebox{\textwidth}{!}{%
\begin{tabular}{r|c|ccc|ccc|ccc|ccc|ccc|ccc}
\toprule
\multirow{2}{*}{\textbf{Methods}} &\multirow{2}{*}{\textbf{Pub.}}
&\multicolumn{3}{c}{\textbf{DUTS-TE}}  
&\multicolumn{3}{c}{\textbf{ECSSD}}  
&\multicolumn{3}{c}{\textbf{DUT-OMRON}} 
&\multicolumn{3}{c}{\textbf{HKU-IS}} 
&\multicolumn{3}{c}{\textbf{SOD}} 
&\multicolumn{3}{c}{\textbf{PASCAL-S}}\\
\cmidrule(r){3-5} \cmidrule(r){6-8} \cmidrule(r){9-11} \cmidrule(r){12-14} \cmidrule(r){15-17}  \cmidrule(r){18-20} 
& & $S_\mathrm{m} \uparrow$  & $\text{max}F \uparrow $ & $\text{MAE} \downarrow$ 
& $S_\mathrm{m} \uparrow$  & $\text{max}F \uparrow $ & $\text{MAE} \downarrow$ 
& $S_\mathrm{m} \uparrow$  & $\text{max}F \uparrow $ & $\text{MAE} \downarrow$ 
& $S_\mathrm{m} \uparrow$  & $\text{max}F \uparrow $ & $\text{MAE} \downarrow$
& $S_\mathrm{m} \uparrow$  & $\text{max}F \uparrow $ & $\text{MAE} \downarrow$  
& $S_\mathrm{m} \uparrow$  & $\text{max}F \uparrow $ & $\text{MAE} \downarrow$  \\
\hline
\hline

\rowcolor[RGB]{220,235,247}FC-SOD~\cite{zhang2020few}&~$\text{NeurIPS}_{20}$  
                            & -   & .846 & .045 
                            & -   & -    & -
                            & -   & .767 & .067
                            & -   & -    & - 
                            & -   & .846 & .122 
                            & -   & .848 & .067
                            \\

LDF~\cite{wei2020label} &~$\text{CVPR}_{20}$    
                            & -   & -    & .034
                            & -   & -    & .034
                            & -   & -    & .051
                            & -   & -    & .027 
                            & -   & -    & -
                            & -   & -    & .060 
                            \\  

\rowcolor[RGB]{220,235,247}MINet~\cite{MINet-CVPR2020}&~$\text{CVPR}_{20}$ 
                            & .884 & .884 & .037
                            & .925 & .947 & .033
                            & .833 & .810 & .055
                            & .920 & .935 & .028
                            & -    & -    & -
                            & .857 & .882 & .064 
                            \\ 

F3Net~\cite{F3Net}&~$\text{AAAI}_{20}$          
                            & .888 & -    & .035
                            & .924 & -    & .033
                            & .838 & -    & .053
                            & .917 & -    & .028 
                            & -    & -    & -
                            & .855 & -    & .062 
                            \\ 

\rowcolor[RGB]{220,235,247}PFPN~\cite{2020Progressive}&~$\text{AAAI}_{20}$ 
                            & .887 & .888 & .037
                            & .932 & .949 & .033
                            & .842 & .820 & .053
                            & .921 & .939 & .030  
                            & -    & -    & -  
                            & .851 & \textcolor{blue}{.892} & .068 
                            \\                             

HVPNet~\cite{liu2020lightweight}&~$\text{TCYB}_{20}$      
                            & -    & -    & .058
                            & -    & -    & .055
                            & -    & -    & .064
                            & -    & -    & .045
                            & -    & -    & .122
                            & -    & -    & - 
                            \\         
                            
\rowcolor[RGB]{220,235,247}CTDNet~\cite{zhao2021complementary}&~$\text{ACM MM}_{21}$ 
                            & -    & -    & .034
                            & -    & -    & .032
                            & -    & -    & .052
                            & -    & -    & .027
                            & -    & -    & - 
                            & -    & -    & .061
                            \\  
                            
MFNet~\cite{piao2021mfnet}&~$\text{ICCV}_{21}$  
                            & .775 & -    & .076
                            & .834 & -    & .084 
                            & .742 & -    & .087
                            & .846 & -    & .059 
                            & -    & -    & -
                            & .770 & -    & .115  
                            \\ 
                            
\rowcolor[RGB]{220,235,247}PFSNet~\cite{ma2021pyramidal} &~$\text{AAAI}_{21}$   
                            & -    & .898 & .036
                            & -    & .952 & .031
                            & -    & .823 & .055
                            & -    & .943 & .026
                            & -    & -    & -
                            & -    & .881 & .063
                            \\  
                            
{SGL-KRN}~\cite{xu2021locate}&~$\text{AAAI}_{21}$ 
                            & -    & .898 & .034
                            & -    & .946 & .036                      
                            & -    & .827 & .049
                            & -    & .939 & .028
                            & -    & -    & - 
                            & -    & .872 & .067 
                            \\

\rowcolor[RGB]{220,235,247}PSGLoss~\cite{yang2021progressive}&~$\text{TIP}_{21}$ 
                            & -    & .890 & .036
                            & -    & .946 & .035                     
                            & -    & .828 & .053
                            & -    & .938 & .027
                            & -    & .872 & .096 
                            & -    & .879 & .061 
                            \\                        
                            
SAMNet~\cite{liu2021samnet}&~$\text{TIP}_{21}$   
                            & -    & .835 & .058
                            & -    & .925 & .053
                            & -    & .797 & .065
                            & -    & .915 & .045
                            & -    & .833 & .123
                            & -    & -    & -
                            \\ 

\rowcolor[RGB]{220,235,247}DPNet~\cite{DPNet_TIP2022} &~$\text{TIP}_{22}$       
                            & .912 & -    & .029
                            & .937 & -    & .028
                            & .853 & -    & .049
                            & .934 & -    & .023
                            & -    & -    & -  
                            & .856 & -    & .072 
                            \\ 

EDN-R~\cite{wu2022edn} &~$\text{TIP}_{22}$       
                            & .892 & .893 & .035
                            & .927 & .950 & .033
                            & .849 & .821 & .050
                            & .924 & .940 & .027
                            & -    & -    & -  
                            & .865 & .879 & .062 
                            \\ 

\rowcolor[RGB]{220,235,247}CSNet~\cite{21PAMI-Sal100K}&~$\text{TPAMI}_{22}$  
                            & -    & .819 & .074
                            & -    & .916 & .066
                            & -    & .792 & .080
                            & -    & .899 & .059
                            & -    & .825 & .137
                            & -    & .835 & .102
                            \\  

TSPORTNet~\cite{TSPORTNet_TPAMI2022} &~$\text{TPAMI}_{22}$       
                            & .871 & -    & .043
                            & .913 & -    & .041
                            & .823 & -    & .058
                            & .909 & -    & .032
                            & -    & -    & -  
                            & .850 & -    & .071 
                            \\ 

\rowcolor[RGB]{220,235,247}DCENet~\cite{DCENet_TCSVT2022} &~$\text{TCSVT}_{22}$       
                            & -    & .889 & .038
                            & -    & .948 & .035
                            & -    & .837 & .055
                            & -    & .935 & .030
                            & -    & \textcolor{red}{.883} & .089  
                            & -    & .885 & .062 
                            \\ 

DHNet~\cite{DHNet_TCSVT2022} &~$\text{TCSVT}_{22}$       
                            & .897 & .886 & .033
                            & .931 & .945 & .031
                            & .850 & .799 & .052
                            & .924 & .935 & .026
                            & -    & -    & -  
                            & .870 & .867 & .059 
                            \\                              

\rowcolor[RGB]{220,235,247}MENet~\cite{MENet_CVPR2023}&~$\text{CVPR}_{23}$   
                            & .905 & .912 & .028
                            & .928 & .955 & .031
                            & .850 & .834 & .045
                            & .927 & .948 & .023
                            & .809 & .878 & .087  
                            & .872 & .890 & .054 
                            \\ 

BBRF~\cite{ma2023boosting}&~$\text{TIP}_{23}$   
                            & .908 & .916 & \textcolor{cyan}{.025}
                            & .939 & \textcolor{cyan}{.963} & \textcolor{cyan}{.022}
                            & .855 & \textcolor{blue}{.843} & \textcolor{blue}{.042}
                            & .935 & \textcolor{red}{.958} & \textcolor{cyan}{.020}
                            & -    & -    & -  
                            & .871 & \textcolor{cyan}{.891} & \textcolor{cyan}{.049} 
                            \\ 

\rowcolor[RGB]{220,235,247}ICON-S~\cite{ICON}&~$\text{TPAMI}_{23}$   
                            & {.917} & -    & \textcolor{cyan}{.025}
                            & {.941} & -    & {.023}
                            & .869 & -    & \textcolor{cyan}{.043}
                            & .935 & -    & .022
                            & .825 & -    & {.083}  
                            & .885 & -    & \textcolor{blue}{.048} 
                            \\ 

PRNet-R~\cite{PRNet_TMM2023}&~$\text{TMM}_{23}$   
                            & .897 & .881 & .039
                            & .917 & .939 & .039
                            & .829 & .804 & .056
                            & .913 & .931 & .032
                            & -    & -    & -  
                            & .853 & .877 & .067 
                            \\ 

\rowcolor[RGB]{220,235,247}VSCode-S~\cite{luo2024vscode}&~$\text{CVPR}_{24}$   
                            & \textcolor{blue}{.926} & \textcolor{cyan}{.922} & -
                            & \textcolor{blue}{.949} & .959 & -
                            & \textcolor{blue}{.877} & \textcolor{cyan}{.840} & -
                            & \textcolor{cyan}{.940} & .951 & -
                            & \textcolor{red}{.870} & \textcolor{blue}{.882} & -  
                            & \textcolor{cyan}{.887} & .864 & -
                            \\ 

VST-S++~\cite{liu2024vst++}&~$\text{TPAMI}_{24}$   
                            & .909 & .897 & .029
                            & .939 & .951 & .027
                            & .859 & .813 & .050
                            & .932 & .941 & .025
                            & \textcolor{cyan}{.859} & .866 & \textcolor{red}{.059}  
                            & .880 & .859 & .062 
                            \\ 
                              
\rowcolor[RGB]{220,235,247}ELSANet~\cite{ELSANet_TCSVT2024}&~$\text{TCSVT}_{24}$   
                            & -    & .882 & .034
                            & -    & .943 & .030
                            & -    & .794 & .050
                            & -    & .935 & .025
                            & -    & -    & -  
                            & -    & .862 & .059 
                            \\ 

ABiUNet~\cite{qiu2024abiunet}&~$\text{TCSVT}_{24}$   
                            & .904 & .906 & .029
                            & .936 & .959 & .026
                            & .860 & \textcolor{blue}{.843} & \textcolor{cyan}{.043}
                            & .932 & .951 & {.021}
                            & .797 & \textcolor{cyan}{.879} & .089  
                            & -    & -    & -
                            \\ 

\rowcolor[RGB]{220,235,247}SENet~\cite{SENet_TIP2025}&~$\text{TIP}_{25}$      
                            & \textcolor{blue}{.926} & \textcolor{blue}{.925} & \textcolor{red}{.022}
                            & \textcolor{blue}{.949} & \textcolor{blue}{.964} & \textcolor{red}{.020}
                            & \textcolor{cyan}{.876} & .835 & \textcolor{red}{.040}
                            & \textcolor{blue}{.941} & \textcolor{cyan}{.953} & \textcolor{blue}{.019}
                            & -    & -    & - 
                            & \textcolor{blue}{.890} & .886 & \textcolor{red}{.046} 
                            \\ 

SDNet~\cite{su2025rapid}&~$\text{TPAMI}_{25}$      
                            & .839 & .816 & .057
                            & .908 & .922 & .045
                            & .818 & .773 & .067
                            & .903 & .915 & .038
                            & .788 & .798 & .113 
                            & .820 & .815 & .089 
                            \\ 

\rowcolor[RGB]{220,235,247}iGAN~\cite{mao2025igan}&~$\text{TCSVT}_{25}$      
                            & .912 & -    & .026
                            & {.941} & -    & .025
                            & .861 & -    & .047
                            & .929 & -    & .023
                            & \textcolor{blue}{.861} & -    & \textcolor{blue}{.060} 
                            & .879 & -    & .053 
                            \\ 

SAMBA~\cite{he2025samba} &~$\text{CVPR}_{25}$     
                            & \textcolor{red}{.932} & \textcolor{red}{.930} & -
                            & \textcolor{red}{.953} & \textcolor{red}{.965} & -
                            & \textcolor{red}{.889} & \textcolor{red}{.859} & -
                            & \textcolor{red}{.945} & \textcolor{blue}{.956} & -
                            & -    & -    & -  
                            & \textcolor{red}{.892} & \textcolor{red}{.896} & -   
                            \\ 

\hline
\rowcolor[gray]{0.85}\textbf{{SASC-USOD}}      
                            & -    & \textcolor{cyan}{.922} & .916 & \textcolor{blue}{.023}   
                            & \textcolor{cyan}{.944} & .957 & \textcolor{blue}{.021} 
                            & .873 & .837 & .044 
                            & \textcolor{cyan}{.940} & .950 & \textcolor{red}{.018}  
                            & .826 & .860 & \textcolor{cyan}{.082}
                            & \textcolor{cyan}{.887} & .889 & \textcolor{red}{.046}
                            \\
\bottomrule
\end{tabular}}
\end{table*}

\subsection{Generalization to Terrestrial Environments}
Although SASC-USOD is specifically designed for underwater environments,
the structural principles explored in this work, including boundary
sensitivity and region-level coherence, are also fundamental to general SOD.
Therefore, the cross-domain generalization of
SASC-USOD is further investigated by directly evaluating SASC-USOD-S on six widely used RGB SOD
benchmarks without additional fine-tuning, including DUTS-TE~\cite{8099887},
DUT-OMRON~\cite{yang2013saliency},
HKU-IS~\cite{li2015visual},
ECSSD~\cite{yan2013hierarchical},
SOD~\cite{movahedi2010design}, and
PASCAL-S~\cite{6909437}.
SASC-USOD is compared with 30 representative SOTA SOD
methods, and the quantitative results are reported in
Table~\ref{tab:USOD2SOD}.
Although SASC-USOD is trained exclusively on underwater data, it achieves
competitive performance across various terrestrial SOD benchmarks.
In terms of MAE, SASC-USOD obtains the best results on HKU-IS and
PASCAL-S, the second-best results on DUTS-TE and ECSSD, and the
third-best result on SOD.
On DUT-OMRON, SASC-USOD achieves an MAE of 0.044, which remains close to
the leading methods.
These results indicate that the structural representations learned by
SASC-USOD can effectively preserve pixel-level prediction accuracy when
transferred across domains.
SASC-USOD also demonstrates strong structural perception capability,
achieving top-three $S_{\mathrm{m}}$ results on DUTS-TE, ECSSD, HKU-IS,
and PASCAL-S.
Nevertheless, SASC-USOD does not consistently achieve the best
performance in max$F$, particularly on DUT-OMRON and SOD.
This limitation is expected because SASC-USOD is optimized for underwater
scenarios, while existing terrestrial SOD methods are specifically
trained and designed for natural image distributions.
Overall, these results demonstrate that the proposed spatially adaptive
structural coordination mechanism retains a degree of cross-domain
generalization beyond underwater environments.

\subsection{Ablation Study}
To validate the effectiveness of the proposed SASC-USOD framework and
analyze the contribution of each component, ablation studies
are conducted.
Specifically, the ablation studies aim to answer the following four
questions:

(1) whether the BS and RC
representations provide complementary structural information for improved RGB-based
USOD?

(2) whether spatially adaptive structural coordination through the SCM
provides advantages over fixed or globally shared fusion strategies?

(3) whether cooperative structural supervision improves structural
representation learning and stabilizes the coordination process?

(4) whether progressive prediction refinement improves the final saliency
prediction quality?

\subsubsection{Effectiveness of Complementary Structural Representations}
\begin{figure}[t]
\centering
\includegraphics[width=0.98\linewidth]{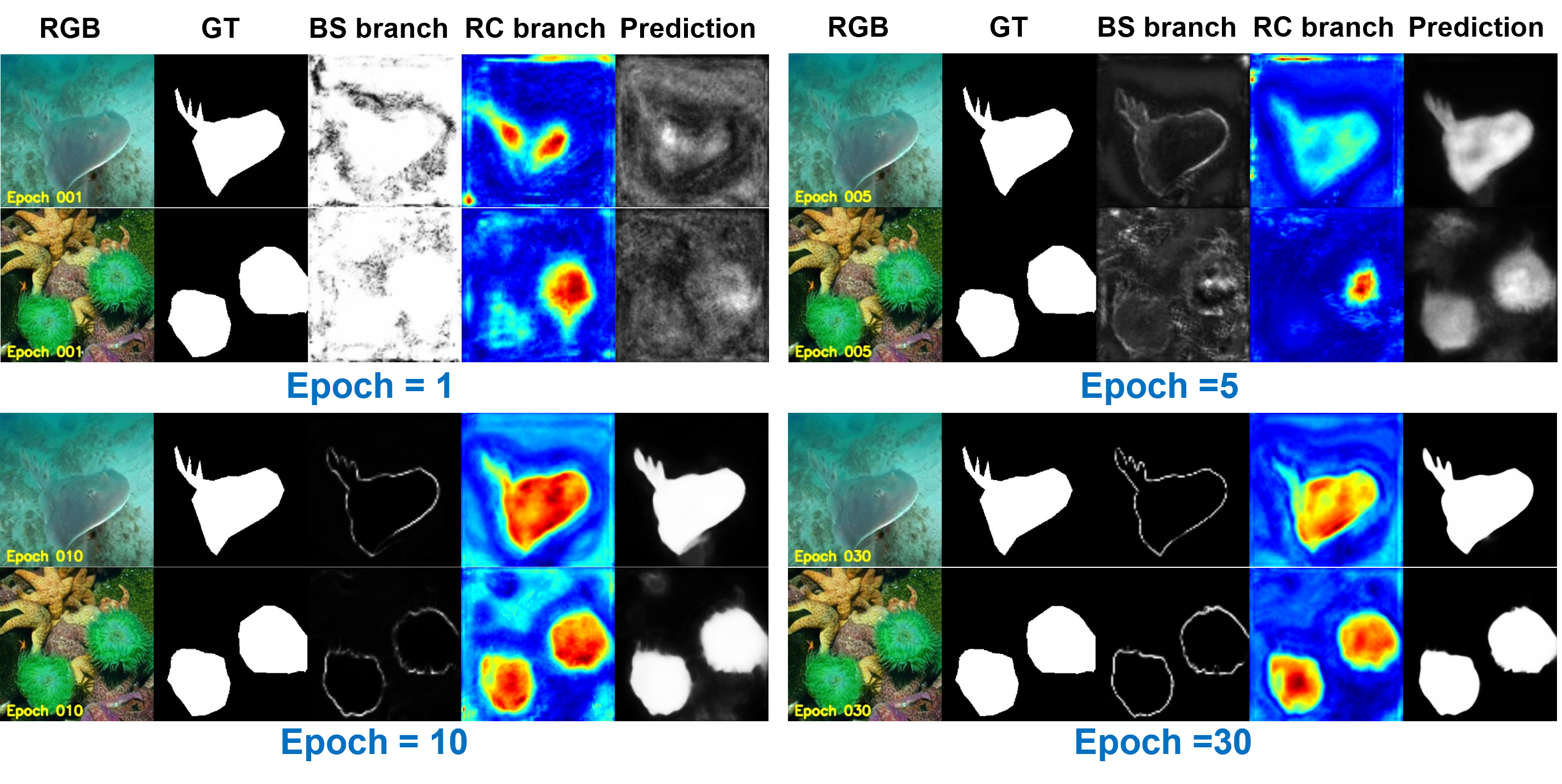}
\caption{
Visualization of the training evolution of the two specialized paths. 
As training proceeds, the BS path gradually learns to emphasize boundaries, while the RC path progressively captures coherent salient regions, leading to increasingly accurate saliency masks.
}
\label{fig:branchvis}
\end{figure}

\textbf{To answer question (1)}, several SASC-USOD variants are constructed
by progressively introducing the BS and RC structural representations.
The baseline directly predicts saliency maps from $\mathbf{F}_{\mathrm{b}}$
without dedicated structural modeling.
The ``+BS'' and ``+RC'' variants introduce only the boundary-sensitive
and region-coherent representations, respectively, while the full
SASC-USOD incorporates both.
As shown in Table~\ref{tab:ablation_dss}, introducing either structural
representation improves the baseline performance on most evaluation
metrics.
The ``+BS'' variant improves $S_{\mathrm{m}}$, max$F$, and MAE, while the
``+RC'' variant increases $S_{\mathrm{m}}$ by 0.41\% and reduces MAE by
1.60\%.
More importantly, incorporating both structural representations achieves
the best performance across all four metrics.
Compared with the baseline, the full SASC-USOD improves
$S_{\mathrm{m}}$ by 1.03\% and max$F$ by 1.33\%, while reducing MAE by
12.23\%.
These results demonstrate that BS and RC
representations provide complementary structural evidence, and that
jointly modeling both structural characteristics is more effective than
using either representation independently.
Fig.~\ref{fig:branchvis} further visualizes the evolution of the two
representations during training.
As training proceeds, the BS representation exhibits increasingly
boundary-focused responses, whereas the RC representation shows more
spatially coherent responses over salient regions.
These response patterns provide additional qualitative
evidence for the complementarity of the two structural representations.

\begin{table}[t]
\caption{Ablation study of structural specialization on USOD10K. ``BS'' and ``RC'' denote the boundary-sensitive branch and the region-coherent branch.}
\label{tab:ablation_dss}
\centering 
\resizebox{\columnwidth}{!}{%
\begin{tabular}{c|cc|cccc} 
\toprule 
\rowcolor[RGB]{220,235,247} Variant & BS & RC & $S_\mathrm{m} \uparrow$ & $E_{\xi}^{\max} \uparrow$ & $\text{max}F \uparrow$ & $\mathrm{MAE} \downarrow$ \\ 
\midrule 
Baseline                & - & -  &.9243 &.9709 &.9241 &.0188 \\ 
+ BS & $\checkmark$     & -      &.9265 &.9657 &.9256 &.0183 \\
+ RC & - & $\checkmark$ &.9281 & .9700& .9319 & .0185 \\ 
\rowcolor[gray]{0.85} SASC-USOD & $\checkmark$ & $\checkmark$ & \textbf{.9338}  & \textbf{.9740} &\textbf{.9364} &\textbf{.0165}  \\ 
\bottomrule
\end{tabular}%
}
\end{table}

\begin{figure}[t]
\centering
\includegraphics[width=0.98\linewidth]{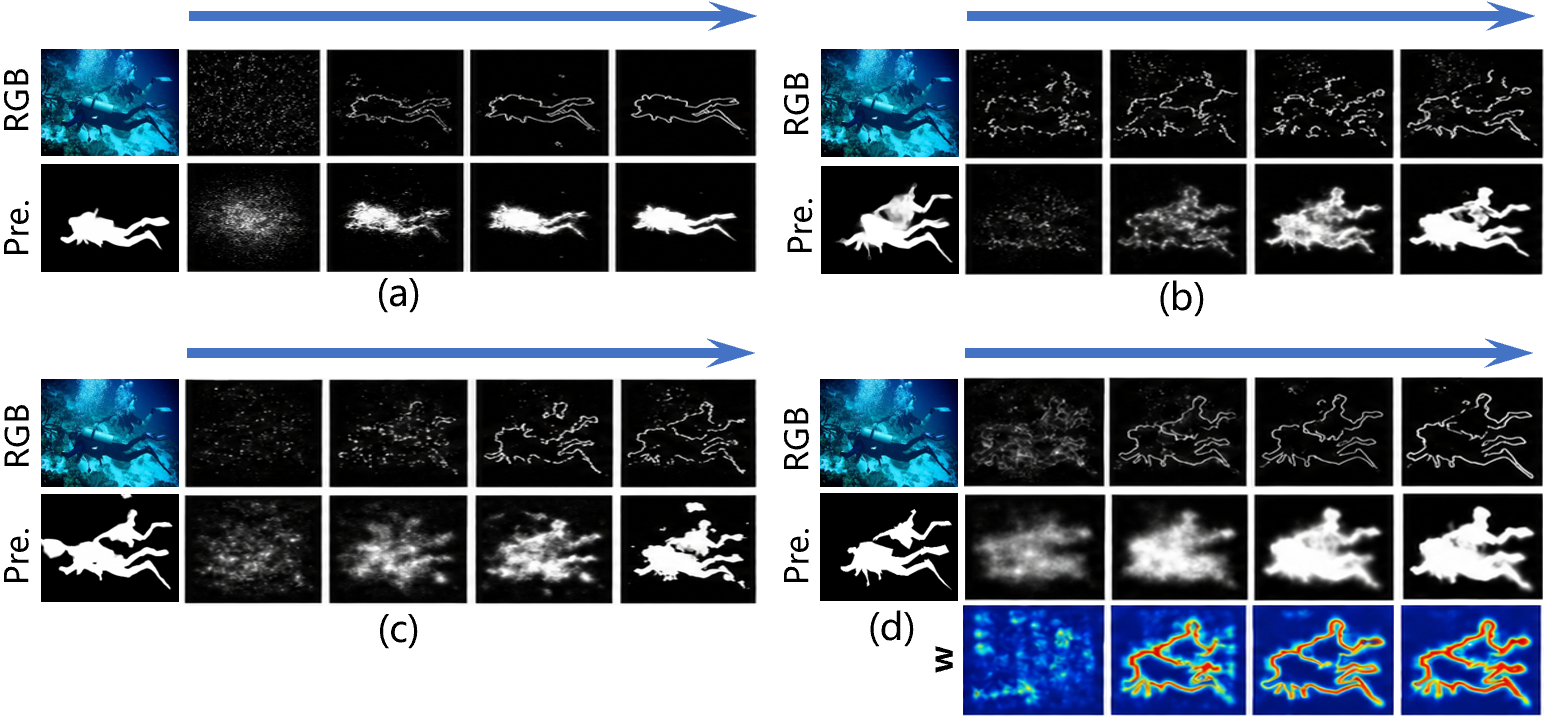}

\caption{
Training progression of BS and RC representations across four methods: (a) BASNet~\cite{qin2019basnet}, (b) EGNet~\cite{9008371}, (c) TC-USOD~\cite{10102831}, and (d) SASC-USOD. 
The first column shows the input image (top) and final saliency mask (bottom), the subsequent columns display intermediate boundary and region feature maps over training. 
SASC-USOD additionally shows weight map ($\mathbf{w}$) of BS learning, showing the effectiveness of the designed SCM.}
\label{fig:boundaryvis}
\end{figure}

To further analyze the difference between SASC-USOD and existing
boundary-aware methods, the evolution of structural responses is
visualized during training in Fig.~\ref{fig:boundaryvis}.
For BASNet~\cite{qin2019basnet}, the boundary predictions are relatively
clear; however, the corresponding saliency maps may still contain
incomplete salient regions in challenging underwater scenes.
EGNet~\cite{9008371} improves salient-object localization by integrating
edge information, but its boundary responses may remain inaccurate,
resulting in relatively coarse object contours.
TC-USOD~\cite{10102831} further improves boundary prediction through
joint saliency and boundary learning, yet some background regions may
still receive incorrect responses under underwater image degradation.
In contrast, SASC-USOD produces cleaner boundary-sensitive responses
while maintaining more coherent region-level responses.
The BS and RC representations exhibit clearly differentiated spatial
patterns during training: the BS representation progressively focuses on
local boundary structures, whereas the RC representation captures
more consistent responses within salient regions.
Moreover, the learned coordination map $\mathbf{w}$ varies across spatial
locations, indicating that SASC-USOD adaptively balances the contributions
of the two structural representations rather than applying a fixed fusion
strategy.
These visualizations provide qualitative evidence that the proposed
spatially adaptive structural coordination effectively preserves boundary
details and region coherence under underwater conditions.

\subsubsection{Effectiveness of Spatially Adaptive Coordination}
\textbf{To answer question (2)}, the proposed SCM is compared with several
alternative structural coordination strategies, including fixed averaging,
learnable global scalar weighting, and concatenation followed by
convolution.
Since the reliability of BS and RC
representations varies across spatial locations under underwater image degradation, fixed or globally shared fusion strategies may
fail to adapt to local structural characteristics.
Therefore, SCM is designed to learn a spatially adaptive coordination
pattern that balances the contributions of the two complementary
structural representations according to image content.

\begin{table}[t]
\caption{Ablation study of different structural coordination strategies.}
\label{tab:ablation_scm}
\centering
\resizebox{\linewidth}{!}{
\begin{tabular}{l|cccc}
\toprule
\rowcolor[RGB]{220,235,247}
Coordination strategy & $S_\mathrm{m} \uparrow$
& $E_{\xi}^{\max} \uparrow$
& $\text{max}F \uparrow$
& $\mathrm{MAE} \downarrow$ \\
\midrule
Fixed averaging           & .9318 & .9724 & .9342 & .0179 \\
Global scalar weighting   & .9321 & .9717 & .9335 & .0173 \\
Concatenation + Conv      & .7125 & .7297 & .7162 & .0510 \\
\rowcolor[gray]{0.85}
SCM                       & \textbf{.9338} & \textbf{.9740}
                          & \textbf{.9364} & \textbf{.0165} \\
\bottomrule
\end{tabular}}
\end{table}

As shown in Table~\ref{tab:ablation_scm}, fixed averaging and global
scalar weighting achieve competitive performance, but remain consistently
inferior to SCM across all four evaluation metrics.
Fixed averaging assigns identical importance to the two structural
representations at all spatial locations, while global scalar weighting
only learns a dataset-level preference and cannot capture local
variations in structural reliability.
In contrast, SCM learns location-dependent structural coordination and
achieves the best performance, improving $S_{\mathrm{m}}$ to 0.9338 and
reducing MAE to 0.0165.
The concatenation-based strategy followed by convolution produces a
substantial performance degradation, suggesting that directly mixing
the two structural representations without explicitly modeling their
spatial relationship is insufficient for degraded underwater scenes.
These results demonstrate that spatially adaptive coordination provides
clear advantages over fixed and globally shared fusion strategies,
confirming the importance of dynamically balancing BS
and RC structural information.

\begin{figure}
\begin{center}
\includegraphics[width=0.98\linewidth]{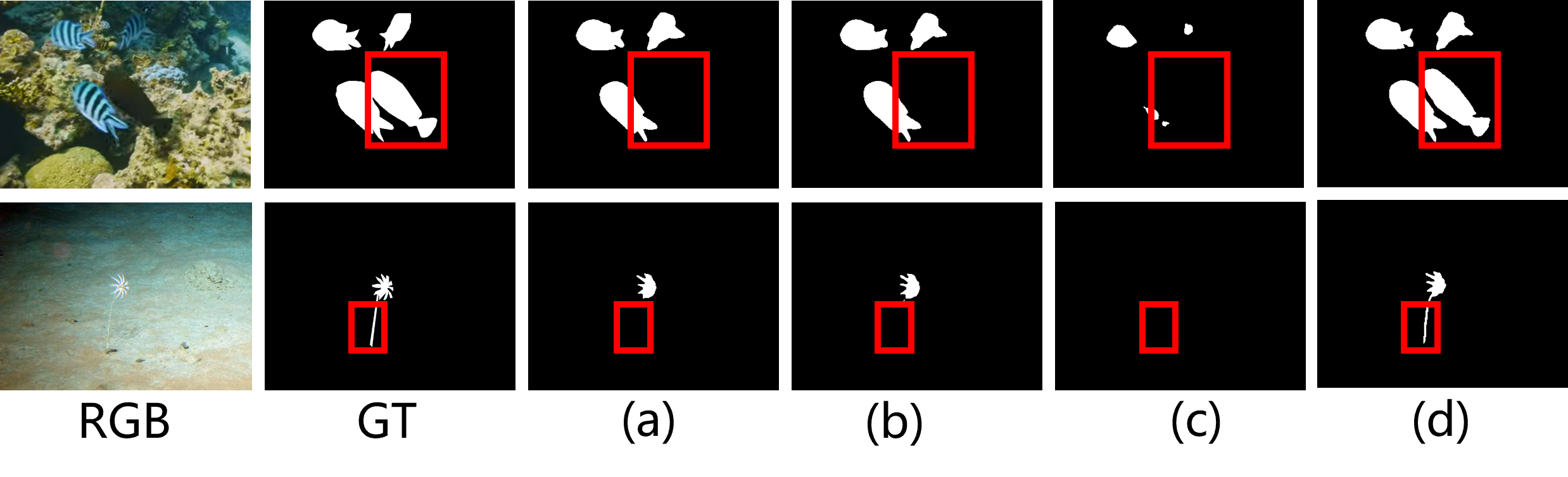}
\end{center}
\caption{
Qualitative comparison of different structural coordination strategies
in SASC-USOD.
(a) Fixed averaging.
(b) Global scalar weighting.
(c) Concatenation + Conv.
(d) SCM.
}
\label{fig:ablationscm}
\end{figure}

Fig.~\ref{fig:ablationscm} further illustrates the differences among
these coordination strategies.
For relatively simple scenes with clear foreground--background
separation, fixed averaging, global scalar weighting, and SCM produce
similar predictions.
However, in challenging cases involving weak boundaries, cluttered
backgrounds, or thin structures, SCM better preserves local structural
details while suppressing distracting background responses.
By contrast, concatenation followed by convolution tends to generate
less accurate predictions, including incomplete salient regions or
imprecise boundaries.
Together with the quantitative results, these observations demonstrate
that SCM can adaptively coordinate complementary structural information
according to local image content.

To investigate the factors influencing spatial coordination,
four SCM variants with different input configurations are compared in
Table~\ref{tab:scm_input}.
Using the shared base representation $\mathbf{F}_{\mathrm{b}}$ as the
baseline input, the predicted boundary response $\hat{\mathbf{E}}$ and
the structural response discrepancy $\mathbf{D}$ are introduced
separately and jointly.
Compared with using only $\mathbf{F}_{\mathrm{b}}$, introducing either
$\hat{\mathbf{E}}$ or $\mathbf{D}$ improves the overall performance,
demonstrating that both boundary-aware prior information and structural
disagreement cues contribute to spatial coordination.
The best performance is obtained when both cues are jointly introduced,
indicating that they provide complementary evidence for learning the
spatial coordination pattern.

\begin{table}[t]
\caption{Ablation study of different input cues for the SCM.}
\label{tab:scm_input}
\centering
\resizebox{\linewidth}{!}{%
\begin{tabular}{l|c|cc|cccc}
\toprule
\rowcolor[RGB]{220,235,247}
Input cues
& $\mathbf{F}_{\mathrm{b}}$
& $\hat{\mathbf{E}}$
& $\mathbf{D}$
& $S_{\mathrm{m}} \uparrow$
& $E_{\xi}^{\max} \uparrow$
& $\mathrm{max}F \uparrow$
& $\mathrm{MAE} \downarrow$ \\
\midrule

$[\mathbf{F}_{\mathrm{b}}]$
& \checkmark
&
&
& .8910
& .9112
& .8913
& .0274 \\

$[\mathbf{F}_{\mathrm{b}},\hat{\mathbf{E}}]$
& \checkmark
& \checkmark
&
& .9247
& .9542
& .9304
& .0203 \\

$[\mathbf{F}_{\mathrm{b}},\mathbf{D}]$
& \checkmark
&
& \checkmark
& .9256
& .9589
& .9283
& .0201 \\

\rowcolor[gray]{0.85}
$[\mathbf{F}_{\mathrm{b}},\hat{\mathbf{E}},\mathbf{D}]$
& \checkmark
& \checkmark
& \checkmark
& \textbf{.9338}
& \textbf{.9740}
& \textbf{.9364}
& \textbf{.0165} \\

\bottomrule
\end{tabular}%
}
\end{table}

\subsubsection{Effectiveness of Cooperative Structural Supervision}
\textbf{To answer question (3)}, several SASC-USOD variants are
constructed by removing boundary supervision, coordination supervision,
or both, to investigate the contribution of cooperative structural
supervision.
The quantitative results are summarized in
Table~\ref{tab:ablation_sup}.
The full model with both supervision terms achieves the best performance
across all four evaluation metrics.
Compared with the variant without auxiliary supervision, the full model
improves $S_{\mathrm{m}}$ from 0.9311 to 0.9338 and max$F$ from 0.9323
to 0.9364, while reducing MAE from 0.0171 to 0.0165.
Removing boundary supervision leads to performance degradation, indicating
that explicit boundary guidance is beneficial for maintaining the
boundary-sensitive characteristics of the BS representation.
Similarly, removing coordination supervision decreases the performance,
demonstrating that additional guidance for SCM is important for learning
effective spatial structural coordination.
These results indicate that the two supervision terms play complementary
roles: boundary supervision encourages specialized structural responses
in the BS representation, while coordination supervision stabilizes the
adaptive balancing between boundary-sensitive and region-coherent
representations.
Jointly applying the two supervision strategies provides more
effective optimization guidance than using either term independently.

\begin{table}[t]
\caption{Ablation study of cooperative structural supervision.
``Bnd.'' and ``Coord.'' denote boundary supervision and coordination
supervision, respectively.}
\centering
\resizebox{\linewidth}{!}{
\begin{tabular}{c|cc|cccc}
\toprule
\rowcolor[RGB]{220,235,247}
Supervisions & Bnd. & Coord.
& $S_\mathrm{m} \uparrow$
& $E_{\xi}^{\max} \uparrow$
& $\text{max}F \uparrow$
& $\mathrm{MAE} \downarrow$ \\
\midrule
w/o both
& - & -
& .9311 & .9719 & .9323 & .0171 \\

w/o Bnd.
& - & $\checkmark$
& .9304 & .9718 & .9333 & .0190 \\

w/o Coord.
& $\checkmark$ & -
& .9297 & .9707 & .9315 & .0185 \\

\rowcolor[gray]{0.85}
Full model
& $\checkmark$ & $\checkmark$
& \textbf{.9338}
& \textbf{.9740}
& \textbf{.9364}
& \textbf{.0165} \\
\bottomrule
\end{tabular}}
\label{tab:ablation_sup}
\end{table}

The qualitative comparisons in Fig.~\ref{fig:ablationcss} further
illustrate the influence of different supervision settings.
For underwater images with relatively simple backgrounds, all variants
can roughly localize salient objects, while the full model produces more
complete salient regions and better-preserved structural details.
In more challenging cases involving ambiguous boundaries or distracting
background patterns, removing either supervision term may result in
incomplete regions, inaccurate object contours, or increased background
responses.
In contrast, the full cooperative structural supervision setting achieves
more accurate saliency predictions by jointly preserving boundary
sensitivity and regional coherence.
These qualitative results further support the effectiveness of jointly
supervising structural representation learning and spatial coordination.

\subsubsection{Effectiveness of the Cooperative Structural Supervision} 
\textbf{To answer question (3)}, several SASC-USOD variants are
constructed by removing boundary supervision, coordination supervision,
or both, to investigate the contribution of the designed cooperative structural
supervision.
The quantitative results are summarized in
Table~\ref{tab:ablation_sup}.
The full model with both supervision terms achieves the best performance
across all four evaluation metrics.
Compared with the variant without auxiliary supervision, the full model
improves $S_{\mathrm{m}}$ from 0.9311 to 0.9338 and max$F$ from 0.9323
to 0.9364, while reducing MAE from 0.0171 to 0.0165.
Removing boundary supervision leads to performance degradation, indicating
that explicit boundary guidance is beneficial for maintaining the
boundary-sensitive characteristics.
Similarly, removing coordination supervision decreases the performance,
demonstrating that additional guidance for SCM is important for learning
effective spatial structural coordination.
These results indicate that the two supervision terms play complementary
roles: boundary supervision encourages specialized boundary-sensitive structural responses
in the BS path, while coordination supervision stabilizes the
adaptive balancing between BS and RC representations.
Jointly applying the two supervision strategies therefore provides more
effective optimization guidance than using either term independently.

\begin{figure}
\centering
\includegraphics[width=0.98\linewidth]{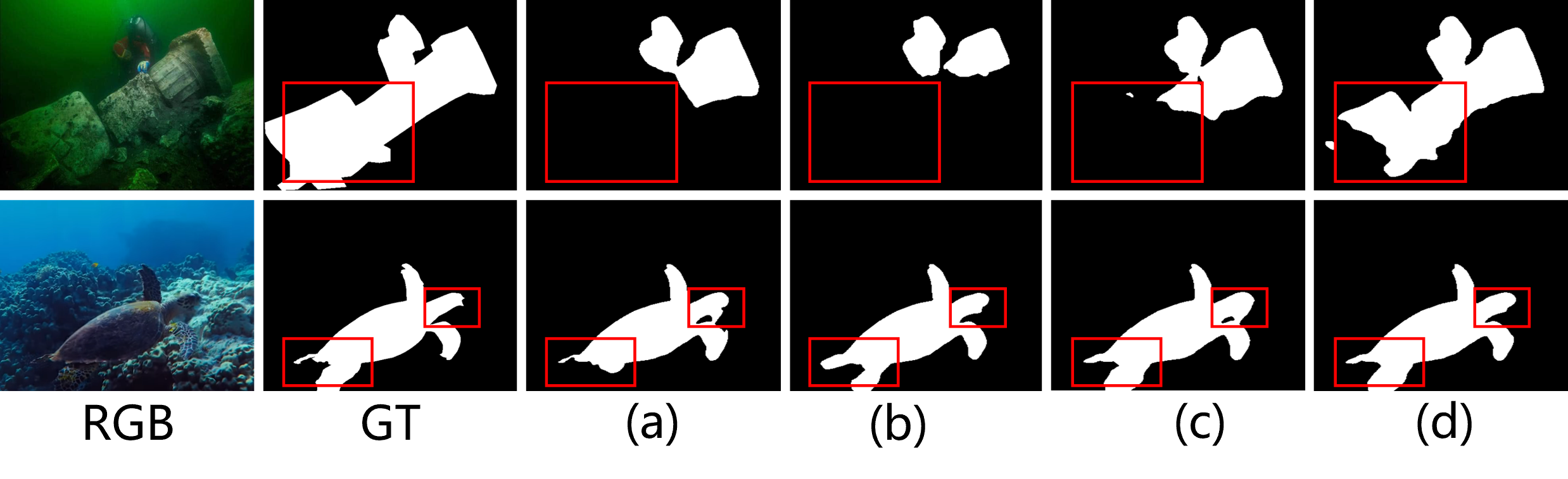}
\caption{
Qualitative comparison of different supervision strategies in the proposed SASC-USOD method. 
(a) without both boundary supervision and coordination supervision. 
(b) without boundary supervision. 
(c) without coordination supervision. 
(d) with full cooperative structural supervision. 
}
\label{fig:ablationcss}
\end{figure}

The qualitative comparisons in Fig.~\ref{fig:ablationcss} further
illustrate the influence of different supervision settings.
For underwater images with relatively simple backgrounds, all variants
can roughly localize salient objects, while the full model produces more
complete salient regions and better-preserved structural details.
In more challenging cases involving ambiguous boundaries or distracting
background patterns, removing either supervision term may result in
incomplete regions, inaccurate object contours, or increased background
responses.
In contrast, the full cooperative structural supervision setting achieves
more accurate saliency predictions by jointly preserving boundary
sensitivity and regional coherence.
These qualitative results further support the effectiveness of jointly
supervising structural representation learning and spatial coordination.

\subsubsection{Effectiveness of Progressive Prediction Refinement}
\textbf{To answer question (4)}, the contribution of the PPR strategy after structural coordination is evaluated.
Two variants of SASC-USOD are constructed for comparison.
SASC-USOD-v1 uses only the low-resolution prediction, which is bilinearly
upsampled to the input resolution as the final output.
SASC-USOD-v2 further introduces the coarse full-resolution prediction stage
and directly uses its output as the final prediction.
The complete SASC-USOD additionally incorporates the final refinement
stage to recover fine spatial details.
The quantitative results are shown in
Table~\ref{tab:ablation_decoder}.
\begin{table}[t]
\centering
\caption{
Ablation study of progressive prediction refinement.
``Low-res'', ``Coarse'', and ``Refine'' denote the low-resolution
prediction, coarse full-resolution prediction, and final refinement
stage, respectively.
}
\label{tab:ablation_decoder}
\resizebox{\columnwidth}{!}{%
\begin{tabular}{c|ccc|cccc}
\toprule
\rowcolor[RGB]{220,235,247}
Methods & Low-res & Coarse & Refine
& $S_\mathrm{m} \uparrow$
& $E_{\xi}^{\max} \uparrow$
& $\text{max}F \uparrow$
& $\mathrm{MAE} \downarrow$ \\
\midrule
SASC-USOD-v1
& $\checkmark$ & - & -
& .5659 & .5669 & .5056 & .1220 \\

SASC-USOD-v2
& $\checkmark$ & $\checkmark$ & -
& .9299 & .9717 & .9324 & .0182 \\

\rowcolor[gray]{0.85}
SASC-USOD
& $\checkmark$ & $\checkmark$ & $\checkmark$
& \textbf{.9338}
& \textbf{.9740}
& \textbf{.9364}
& \textbf{.0165} \\
\bottomrule
\end{tabular}%
}
\end{table}
Directly using the
upsampled low-resolution prediction leads to limited accuracy,
achieving an $S_{\mathrm{m}}$ of 0.5659 and an MAE of 0.1220.
Introducing the coarse full-resolution prediction stage substantially
improves performance, increasing $S_{\mathrm{m}}$ to 0.9299 and reducing
MAE to 0.0182.
This indicates that recovering predictions at higher spatial resolution
is essential for accurately localizing salient objects.
The final refinement stage further improves all four evaluation metrics,
achieving 0.9338 in $S_{\mathrm{m}}$, 0.9740 in
$E_{\xi}^{\max}$, 0.9364 in max$F$, and 0.0165 in MAE.
These improvements demonstrate that PPR strategy effectively
recovers fine spatial details after structural coordination, leading to
more accurate saliency map prediction.

\begin{figure}[t]
\centering
\includegraphics[width=0.98\linewidth]{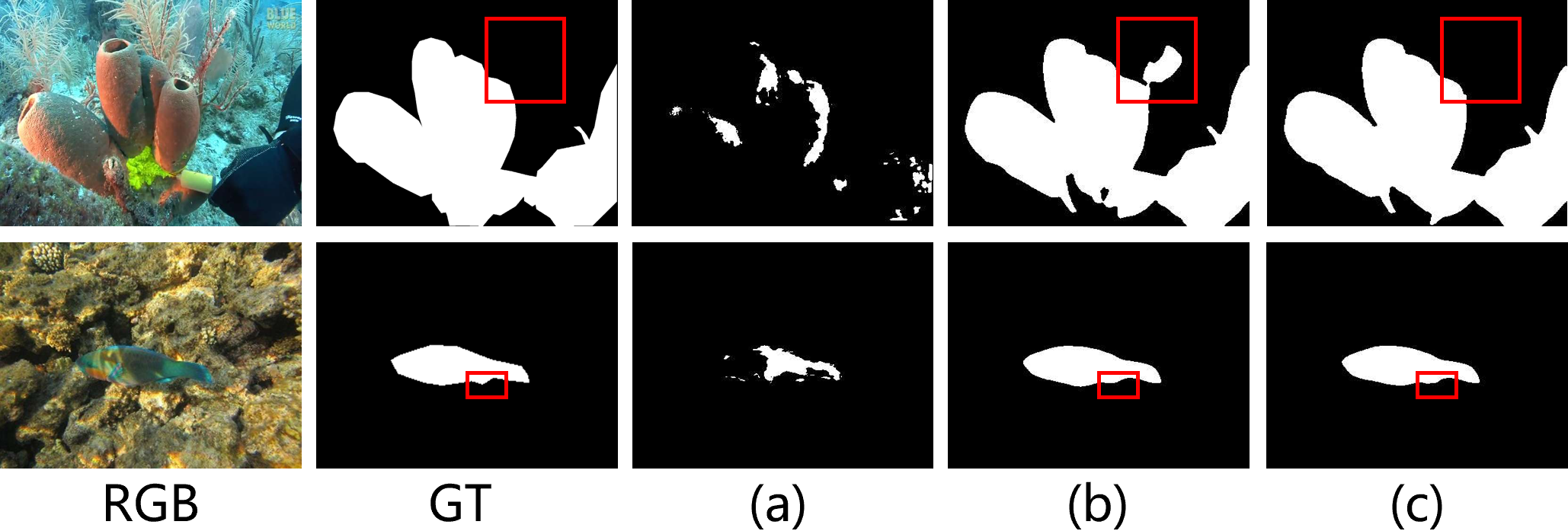}

\caption{
Qualitative comparison of different PPR variants. 
From left to right: input image, ground truth, (a) SASC-USOD-v1 prediction, (b) SASC-USOD-v2 prediction, and (c) SASC-USOD prediction. 
}
\label{fig:pprablation}
\end{figure}

The qualitative comparisons in Fig.~\ref{fig:pprablation} further
illustrate the effect of PPR strategy.
SASC-USOD-v1 can roughly localize salient objects but produces incomplete
regions and coarse boundaries due to the limited spatial resolution.
SASC-USOD-v2 generates more complete salient regions after introducing
the coarse full-resolution prediction stage, although some inaccurate
responses remain.
In contrast, the complete SASC-USOD produces cleaner saliency maps with
better-preserved boundaries and fewer distracting background responses.
Together with the quantitative results, these observations demonstrate
that the PPR strategy effectively improves spatial
detail recovery and complements the preceding structural coordination
process.

\subsection{Real-world Application}
\begin{figure}[t]
\centering
\includegraphics[width=0.98\linewidth]{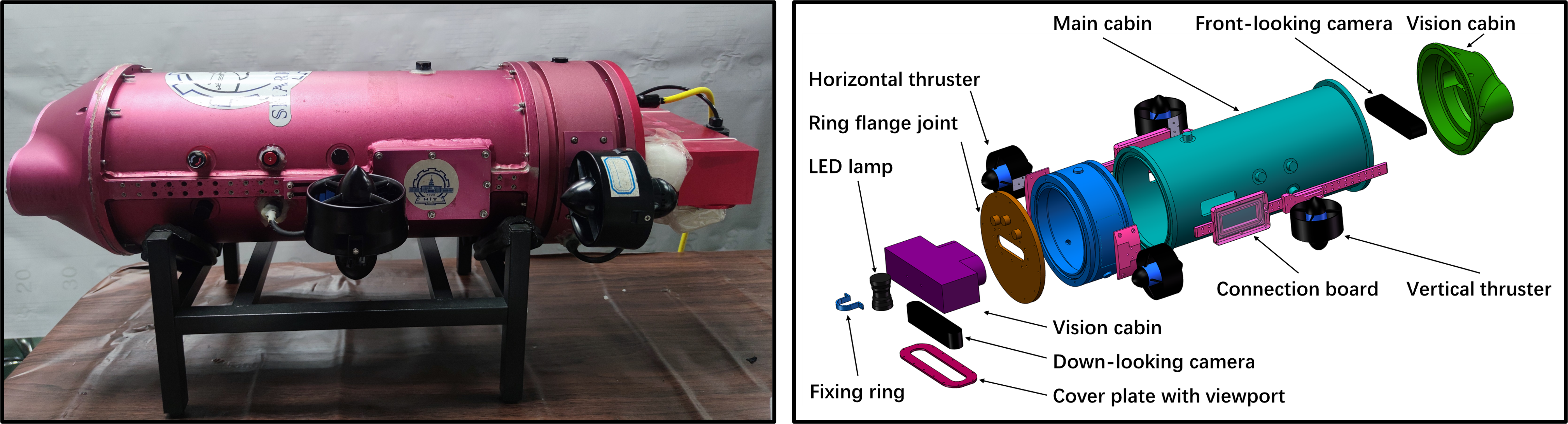}
\caption{Underwater robot used for onboard deployment of
SASC-USOD.}
\label{fig:AUV}
\end{figure}

To evaluate the onboard deployment capability of SASC-USOD,
the trained SASC-USOD-T model is deployed on an underwater robotic platform and
conduct a close-range underwater object inspection experiment.
As shown in Fig.~\ref{fig:AUV}, the robot is equipped with RGB cameras
and an onboard embedded computing unit.
The perception system runs on an NVIDIA Jetson TX2 NX module, where
the captured RGB images are processed locally without relying on an
external workstation.
During deployment, the input images are resized to $352\times352$,
consistent with the training and evaluation settings.
SASC-USOD-T operates in a fully feed-forward manner and directly
generates saliency masks from single underwater images.

\begin{figure}[t]
\centering
\includegraphics[width=0.98\linewidth]{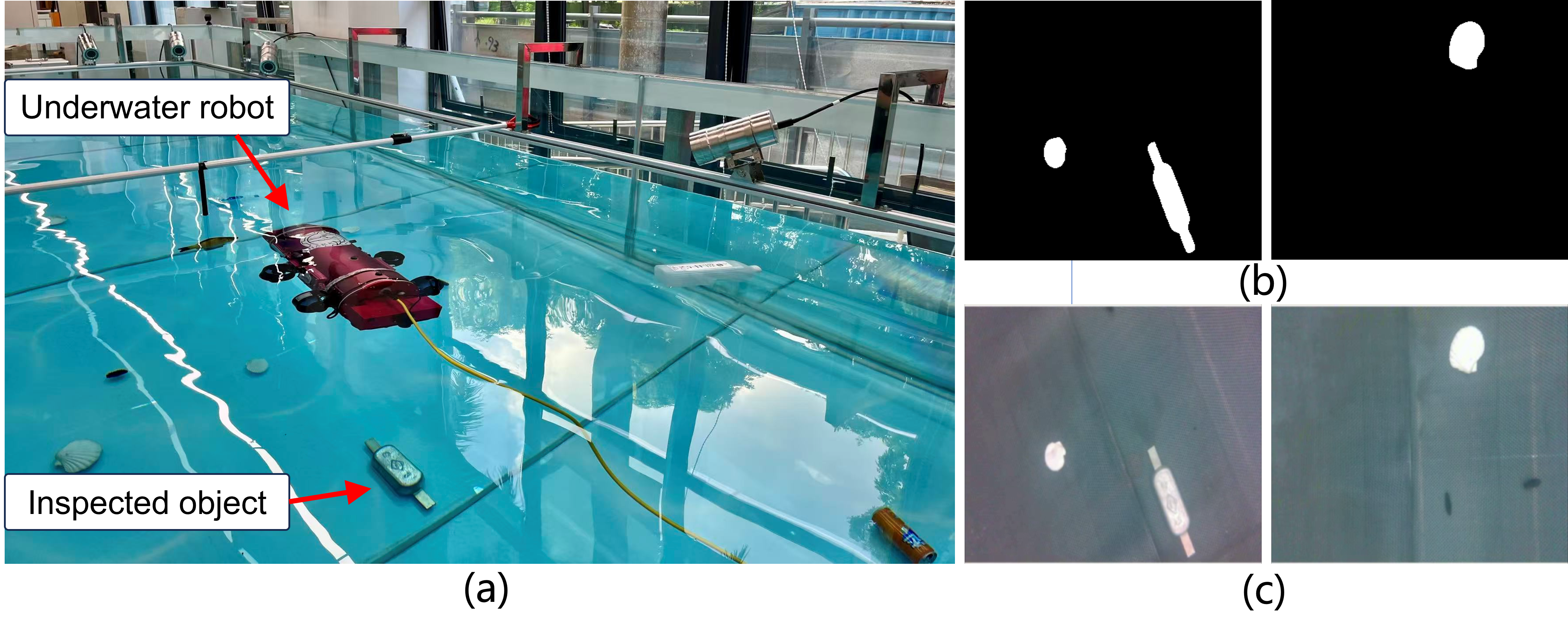}
\caption{Underwater robotic visual target inspection.
(a) Experimental setup.
(b) Camera view of the target.
(c) Saliency masks generated by SASC-USOD.}
\label{fig:realworld_task1}
\end{figure}

Fig.~\ref{fig:realworld_task1}(a) shows the experimental setup,
including the water tank, target object, and underwater robot.
During deployment, SASC-USOD-T requires no auxiliary sensing modality,
iterative optimization, or task-specific post-processing.
On the Jetson TX2 NX, SASC-USOD-T achieves an inference speed
of 21 FPS, demonstrating onboard inference capability on a
resource-constrained underwater robotic platform.
During the experiment, the robot performs close-range visual
observation of underwater objects while continuously capturing RGB
images.
As illustrated in Fig.~\ref{fig:realworld_task1}(b) and
Fig.~\ref{fig:realworld_task1}(c), the predicted saliency maps
successfully highlight visually distinctive target regions while
suppressing distracting background responses in the shown examples.
Such predictions can provide foreground localization cues for
subsequent vision-guided operations, such as region-of-interest
selection, object-centered observation, and visual inspection.
These experiments demonstrate the feasibility of deploying SASC-USOD on an underwater
robot for underwater robotic applications. The video is available at:~\url{https://github.com/LinHong-HIT/SASC-USOD}.

\section{Conclusion}
\label{sec:con}
This paper investigated how complementary structural cues can be
effectively coordinated for RGB-based USOD.
To address the spatial variation of structural cue reliability under
underwater degradation, this paper proposed SASC-USOD, which constructs
boundary-sensitive and region-coherent representations from a shared
feature representation and learns their spatially adaptive coordination.
Unlike spatially uniform fusion strategies, SASC-USOD adaptively
balances the contributions of complementary structural representations
according to local image content.
Cooperative structural supervision further preserves differentiated
structural behaviors and stabilizes the coordination process, while
progressive prediction refinement improves the recovery of fine spatial
details in the final saliency prediction.
Extensive experiments on the USOD10K and USOD benchmarks demonstrate that
SASC-USOD achieves superior performance, reducing MAE by 4.07\% and
23.53\% compared with the strongest competing methods on the two
benchmarks, respectively.
The lightweight SASC-USOD-T variant further achieves 21 FPS on an
NVIDIA Jetson TX2 NX, demonstrating its capability for onboard
underwater robotic perception.
Overall, this work demonstrates the effectiveness of spatially adaptive
coordination between complementary structural representations and
provides a promising solution for RGB-based USOD.

\section*{Acknowledgment}
The work described in this paper was partially supported by grants AoE/E-601/24-N, 16203223, C6029-23G, and C6078-25G from the Research Grants Council of the Hong Kong Special Administrative Region, China.
The authors would like to express their sincere gratitude to Prof. Deng-Ping Fan for his valuable suggestions and insightful discussions.

\bibliography{main}
\bibliographystyle{IEEEtran}

\end{document}